\documentclass[11pt,a4paper]{article}
\usepackage{times,latexsym}
\usepackage{latexsym}
\usepackage{booktabs}
\usepackage{CJKutf8} 
\usepackage{url} 
\usepackage{amsmath}
\usepackage{pgfmath}
\usepackage{float}
\usepackage{graphicx}
\usepackage[table]{xcolor}
\usepackage{xspace}
\usepackage{enumitem}
\usepackage{tabularx}
\usepackage{mfirstuc,tabulary}
\usepackage{colortbl}
\usepackage{pifont}
\usepackage{amssymb}
\usepackage{wasysym}
\usepackage{subcaption}
\usepackage{multirow}
\usepackage{fontawesome5}
\usepackage[ruled,vlined,linesnumbered]{algorithm2e}
\usepackage[most]{tcolorbox}
\usepackage[T1]{fontenc}

\usepackage[]{acl}
\usepackage{microtype}
\definecolor{gray-blue}{HTML}{93BFE7} 
\definecolor{gray-red}{HTML}{F3CCD1} 
\definecolor{blt}{HTML}{3d5a80}
\definecolor{bltl}{HTML}{98c1d9}
\definecolor{orange}{HTML}{DE6B61}
\definecolor{dred}{HTML}{d90429}
\definecolor{tableblue}{HTML}{0072B2}
\definecolor{tableamber}{HTML}{E69F00}
\definecolor{nblue}{HTML}{86C5DA}

\newcommand{\shadeblue}{\edef\cellshade{\noexpand\cellcolor{gray-blue!\shade}}\cellshade}
\newcommand{\shadeamber}{\edef\cellshade{\noexpand\cellcolor{tableamber!\shade}}\cellshade}
\newcommand{\shadeog}{\edef\cellshade{\noexpand\cellcolor{orange!\shade}}\cellshade}

\newcommand{\srepc}[1]{\pgfmathtruncatemacro{\shade}{6+83*(#1+0.23)/(0.34+0.23)}\shadeblue#1}
 
\newcommand{\goalc}[1]{\pgfmathtruncatemacro{\shade}{2+38*(#1+0.4 )/(0.5+0.4)}\shadeog#1}

\newif\iftaclinstructions
\taclinstructionsfalse 
\iftaclinstructions

\newcommand{\instr}
\fi

\newtcolorbox{promptbox}[1][]{
  colback=gray!5!white,
  colframe=gray!75!black,
  arc=3mm,
  boxrule=0.8pt,
  left=8pt,
  right=8pt,
  top=6pt,
  bottom=6pt,
  fonttitle=\bfseries,
  title=Prompt,
  #1
}

\title{Cultural Adaptation within Repeated Social Interactions}

\author{
  Chen Cecilia Liu$^1$
  \and
  Roi Reichart$^2$
  \and
  Anna Korhonen$^1$
  \\\\
  $^1$Language Technology Lab - University of Cambridge \\
  $^2$Technion - Israel Institute of Technology
}

\date{}

\begin{document}
\maketitle
\begin{abstract}
Much of human culture is tacit and is typically learned through situated social interaction. As LLMs are deployed as interactive agents across diverse communities, assessing their cultural competence demands moving beyond existing question-answering approaches, shifting from what they know to how they adapt. In this work, we broaden the scope of cultural adaptation in NLP by investigating whether LLMs can adapt through repeated interactions via a sandbox environment. By reframing cultural adaptation as a continuous process of norm assimilation through social interaction, our work demonstrates that memory mechanisms foster longitudinal learning across repeated interactions, facilitate transfer to different social settings, and enable multilingual norm assimilation across five languages. Together, our findings lay the groundwork for continuous, interaction-driven cultural adaptation in LLMs. 
\end{abstract}

\section{Introduction}
Much of human culture is tacit \cite{polanyi2009tacit} and acquired from situated social interaction \cite{rogoff1990apprenticeship, tomasello2016cultural}. Through repeated encounters and social feedback \cite{bandura1977social, MORRIS20151}, individuals learn to navigate shared norms, expectations, and behavioural conventions \cite{Bicchieri_2005}. As large language models (LLMs) increasingly act as interactive agents across diverse contexts, evaluating their cultural competence requires moving beyond \emph{what they know} to \emph{how they adapt}. This raises three questions: are LLMs capable of assimilating cultural norms through repeated social interaction, sustaining the resulting behavioural adaptations over time, and transferring them to unfamiliar situations?

Existing work in NLP largely treats culture as a static collection of knowledge \cite{zhou-etal-2025-culture}, probed through knowledge- or reasoning-based question answering \cite{chiu-etal-2025-culturalbench, rao-etal-2025-normad, cultureforest/abs-2606-01879}. While not all cultural knowledge requires continuous, interactive adaptation, highly contextual elements like cultural norms naturally demand it. Recent efforts have also shifted toward multi-turn conversations \cite{wu-etal-2025-socialcc, havaldar-etal-2025-culturally, tan-etal-2026-cultureconverse}, yet they typically treat models as fixed entities, measuring immediate behavioural compliance rather than \emph{adaptive plasticity}---the ability to learn from and apply experience in unfamiliar situations. A model that excels at question answering may still fail to acquire implicit cultural norms and expectations within interactions, motivating the evaluation of adaptive mechanisms that help models learn progressively within interactions and social feedback.

In humans, one bridge connecting isolated social interactions to cultural competence is built on \emph{repeated} experiences and supported by memory \cite{strauss1997cognitive, schank2013scripts}. For LLMs, external memory offers a simple but effective way to plasticity without updating model weights. While memory-based agents already use it to maintain behavioural coherence \cite{memory/uist/ParkOCMLB23, DBLP:conf/nips/ShinnCGNY23, DBLP:journals/corr/abs-2312-03664, DBLP:conf/aaai/ZhongGGYW24}, these frameworks primarily assume cultural universality, overlooking the role of memory in the progressive assimilation of implicit norms in repeated interactions.  

To address these gaps, we first reframe cultural adaptation as norm assimilation across repeated social interactions. We instantiate this in a five-culture sandbox environment, where social goal-driven conversation and environmental incentives compel models to recognize and infer implicit norms.
Next, we use the sandbox as a diagnostic testbed to evaluate the norm assimilation capability of state-of-the-art open-source LLMs across three model families, investigating how simple external memory mechanisms\footnote{We focus on external (non-parametric) memory rather than parametric weight fine-tuning for both conceptual and technical reasons (\S\ref{sec:method_memory}).} affect longitudinal learning over repeated interactions. 
Our results suggest that memory mechanisms can generally assimilate cultural norms across both English and multilingual simulations. Furthermore, we find that memory with session-level updates is better at sustaining longitudinal learning than turn-level adaptation, and demonstrates adaptive plasticity with transfer across different scenarios.

In summary, our contributions are: (1) a human- and cognitively inspired framing of culture as a continuous process of norm assimilation and behavioural adaptation through social interactions, grounded in theories of social learning; (2) providing a sandbox environment for evaluating LLMs' cultural adaptation in a dynamic, interactive environment; (3) examining multiple models and memory mechanisms, suggesting that memory helps cultural norm assimilation in both English and multilingual simulations. 

Together, our findings lay the groundwork for a continuous, interaction-driven cultural adaptation.

\section{Related Work}
\noindent\textbf{Cultural Adaptation and Evaluation.} LLMs often reflect cultural values and norms in Western and English-speaking contexts, which are not fully addressed by existing adaptation methods \cite{liu-etal-2025-culturally}. Cultural adaptation has primarily relied on prompting \cite{alkhamissi-etal-2024-investigating, 10.1093/pnasnexus/pgae346} or on training with surveys or narratives \cite{li2024culturepark, adilazuarda-etal-2025-surveys}. For evaluation, benchmarks such as CultureBank \cite{shi-etal-2024-culturebank}, NormAd \cite{rao-etal-2025-normad}, and CultureForest \cite{cultureforest/abs-2606-01879} assess cultural dimensions and local norms primarily in static contexts.

\noindent\textbf{Culture and Norm Assimilation.} 
Static cultural evaluations overlook that culture is largely enacted through interaction, a foundational insight from related disciplines outside of NLP. Rather than following explicit rules, humans infer and internalize implicit social norms \cite{cialdini1990focus, Bicchieri_2005} via repeated exposure, social learning, and reputational feedback \cite{bandura1977social, FEHR2004185, CHUDEK2011218, MORRIS20151}, adapting their behaviour over time \cite{Berry_acculturation, tankard2016norm}. Grounding LLM cultural adaptation in these cognitive mechanisms requires interactive environments where norms must be inferred and assimilated through experience.

\noindent\textbf{Social Simulation, Memory, and Norms.}
Recent studies have instead used LLM-based agents to simulate social interactions and collective behaviours. Building on \textsc{Sotopia}'s \citep{zhou2024sotopia} interactive environment for goal-driven social scenarios, subsequent work explores agent training, large-scale simulation, and extended evaluations \citep{wang-etal-2024-sotopia, zhou-etal-2025-sotopia, tang-etal-2025-gensim}. Concurrently, \citet{pham-etal-2026-liveculturebench} test multi-turn norm robustness under adversarial pressure, which differs fundamentally from our goals. 

Many LLM-based general social simulators also incorporate memory and reflection to support individual perception and planning \cite[inter alia]{shao-etal-2023-character, DBLP:journals/corr/abs-2307-14984} and drive emergent collective dynamics \citep{memory/uist/ParkOCMLB23, DBLP:journals/corr/abs-2502-08691}. In parallel, studies on structured games (e.g., social dilemmas, bargaining) show that environmental incentives, interaction history, and reflection shape behavioural outcomes \citep{liu2025cooperative, DBLP:journals/jair/ShapiraMRT26, he2026social}. 
Together, these studies suggest that emergent normative behaviour depends not only on agent capabilities and prompts, but also on environmental incentives, reflection, and internalization.

While related to \citet{zhou2024sotopia}, our work prioritizes cultural norm \textit{assimilation}, rather than achieving social goals or simulation authenticity. We investigate how external memory mediates norm assimilation across repeated interactions.

\begin{figure*}
    \centering
    \includegraphics[width=0.98\linewidth]{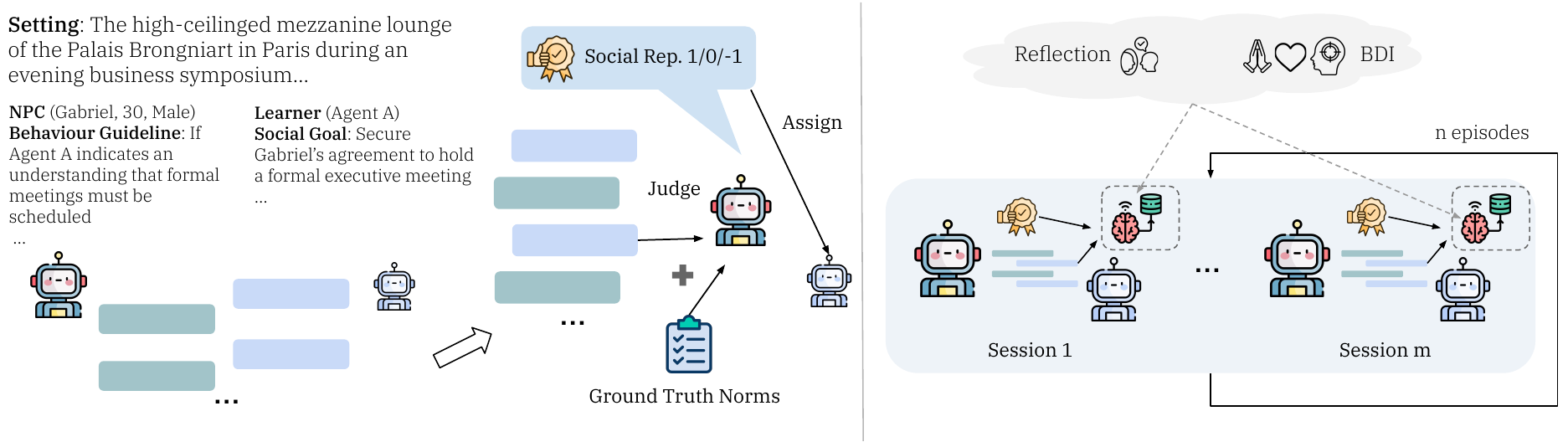}
    \caption{Overview of the setup. Left: An NPC with an assigned cultural identity interacts with an unassigned Learner agent driven by a social goal. To incentivize learning implicit norms, the goal is granted only when the learner demonstrates associated cultural norms and avoids cultural faux pas. A judge model evaluates the dialogue session and assigns a social reputation score in the end. Right: We evaluate how reasoning and memory mechanisms shape longitudinal learning behaviours and norm assimilation across repeated interaction episodes. }
    \label{fig:main}
\end{figure*}

\section{Environment}

Here, we describe a lightweight simulation environment to help evaluate models' adaptation to cultural norms within social interactions. The environment consists of 1,264 scenarios across five distinct cultures, selected for geographical and linguistic diversity: China (114), France (216), Japan (359), New Zealand (333), and Saudi Arabia (242). Their higher-resource status was chosen intentionally to minimize confounding from suboptimal LLM language capabilities.

\subsection{Social Scenarios}
\label{sec:soc_scenarios}
To evaluate models' ability to adapt to cultural norms in social interactions, we set up social interaction scenarios between two agents grounded in real-world cultural norms from CultureForest (\citealt{cultureforest/abs-2606-01879}, norms originally derived from Cultural Atlas; \citealt{sbs_cultural_atlas}) and CultureBank \cite{shi-etal-2024-culturebank}. The scenarios are set up with two requirements: 
\begin{itemize}[noitemsep]
    \item Cultural norms should \textit{unfold naturally during the interaction}, without explicit teaching or explanation.
    \item The social goal should not be granted unless the cultural norms are demonstrated during the interaction. 
\end{itemize}
The first requirement is intended to evaluate whether models can infer and adapt to cultural norms implicitly from the social context. The second requirement establishes an environmental incentive for models to adapt (i.e., goal achievement is the \textit{behavioural motivation or conversation driver} and cultural norm is the gating condition\footnote{Note that the environmental incentive is unidirectional. Cultural norms violation blocks the social goal, but cultural norms compliance does not necessarily guarantee the goal achievement.}), thereby providing implicit signals that push the agent to reflect on and adjust its behaviour.\footnote{We acknowledge that social interactions can be formulated beyond using social goals as a driving motivation. While our work establishes a foundational starting point, future work could extend this paradigm to other open-ended, non-goal-directed interactions and diverse social contexts.}

All scenarios were generated using Gemini-3.7-Flash \cite{deepmind2026gemini37flash} with near gender parity (45.25\% male vs.\ 54.75\% female). Detailed age distributions are provided in Appendix Table~\ref{tab:age_distribution}, while \S\ref{sec:qc} details the scenario quality checks.

Following the existing multi-agent social interaction setup \cite[inter alia]{zhou2024sotopia}, we formalize each social scenario as a structured dyadic environment:
\begin{equation}
\mathcal{S} = \langle \mathcal{E}, \text{NPC}, \mathcal{A} \rangle,
\end{equation}
$\mathcal{E}$ denotes the global sessional environment and contextual setting (details in \S\ref{subsec:formalism}), $\text{NPC}$ represents the non-player character specification, and $\mathcal{A}$ denotes the learner agent. For each interaction session, the environment provides the overarching social setting and contextual constraints.

\noindent\textbf{NPC Agent ($\text{NPC}$).}
The NPC is fully instantiated with a structured profile:
\begin{equation}
\text{NPC} = \langle \text{name}, \text{gender}, \text{rel}(\mathcal{A}), \text{background}, \mathcal{L} \rangle,
\end{equation}
where $\text{name}$ is sampled to be culturally consistent with the setting, $\text{gender}$ and $\text{background}$ provide persona variation, and $\text{rel}(\mathcal{A})$ explicitly defines the social relationship and power dynamic relative to the learner agent $\mathcal{A}$. The NPC is additionally conditioned on a set of scenario-specific behavioural guidelines $\mathcal{L}$, while the ground-truth cultural norms $\mathcal{N}$ remain available during the interaction.

\noindent\textbf{Learner Agent ($\mathcal{A}$).}
To remove persona-induced priors, the evaluated model $\mathcal{A}$ is not conditioned on a predefined persona. Instead, $\mathcal{A}$ is conditioned strictly on its social goal $\mathcal{G}$, requiring the model to uphold contextual norms through dialogue and action within each social interaction.

Table \ref{tab:appendix_scenario_nz} in the appendix provides an example scenario.

\subsection{Interaction Formalism}
\label{subsec:formalism}

We model the multi-session environment as a sequence of discrete interaction sessions $\mathcal{E}_1, \mathcal{E}_2, \dots, \mathcal{E}_T$, where each session $\mathcal{E}_t$ is instantiated with a contextual tuple $(c_t, s_t, g_t, b_t)$.
Here, $c_t \in \mathcal{C}$ is the target cultural group (e.g., Japanese), $s_t \in \mathcal{S}$ is the social setting (e.g., formal business dinner or tea house), $g_t \in \mathcal{G}$ is the social goal of $\mathcal{A}$, and $b_t \in \mathcal{B}$ denotes the other persona and relationship profile of the NPC.

Two agents generate a $K$-turn trajectory per session $\mathcal{E}_t$: $\tau_t = \big( (u_{t,k}, v_{t,k}) \big)_{k=1}^K$, with utterances sampled as $u_{t,k} \sim \pi_{\mathcal{A}}(\cdot \mid \mathcal{H}_{t,k}, \mathcal{M}_t)$ and $v_{t,k} \sim \pi_{\text{NPC}}(\cdot \mid \mathcal{H}_{t,k})$ given history $\mathcal{H}_{t,k}$ and optional memory content $\mathcal{M}_t$.

\subsection{Metric}
\label{subsec:eval}
Because our primary focus is norm assimilation rather than social task competence, we evaluate agents strictly on their adherence to social norms. Social goals serve merely as an elicitation scaffold to motivate agent interaction (\S\ref{sec:soc_scenarios}). Furthermore, because goal attainment can be confounded by non-normative factors (e.g., model persuasion, task complexity, and norm compliance is a necessary but not sufficient condition for goal success), it is neither factored into our evaluation metrics nor used as explicit feedback.

Instead, upon session termination, an LLM judge evaluates the interaction history for cultural-norm alignment to assign a social reputation score, which is used as the signal for learning in our experiment.

\noindent\textbf{Social Reputation Score.} An LLM judge evaluates $\mathcal{A}$'s behaviour against cultural norms required for the scenario, producing a rationale and scalar reputation update $r_t \in \{-1, 0, 1\}$ at each session $t$. A score of 1 indicates that the model complied with the social norm in this scenario.

\subsection{Scenarios Quality Control}\label{sec:qc}
All source norms for scenario construction are acquired from humans from existing resources (\citealt{cultureforest/abs-2606-01879}; \citealt{sbs_cultural_atlas}), making additional validation unnecessary.
To validate the synthetic scenarios, we sampled 100 items (20 per culture, $\sim$10\% of all scenarios) and recruited in-culture evaluators via Prolific to assess scenario acceptability and screen for cultural stereotyping (instructions in Appendix \ref{app:hi}). Overall, 96\% of the sampled items were rated as culturally acceptable and plausible, and 90\% were judged free of stereotyping, with the few flagged cases concentrated primarily in depictions of Japanese female roles. While eliminating biases is essential, this strong baseline confirms that the benchmark provides a reliable testbed for analyzing core norm-assimilation mechanisms. We treat the targeted improvement of scenario generation as an important priority for future work.

\section{Evaluated Methods and Mechanisms}\label{sec:methods}
We evaluate one non-memory (\S\ref{subsec:sr}) and two memory-based methods (\S\ref{sec:method_memory}), with full prompts in Appendix \ref{app:prompts} and pseudocode in Algorithm \ref{alg:memory_bdi_r}.

\subsection{Situational Reasoning}\label{subsec:sr}
Situational Reasoning (SR) is a prompting method in the style of Chain-of-Thought \cite{wei-etal-2022-chain} that encourages explicit, step-by-step reasoning about the interaction, including $\mathcal{A}$'s own goals, the situational context, and applicable cultural norms. This approach is applied at every turn and conditions on the current state of the dialogue, which emphasizes \emph{short-term} social reasoning and rapid adaptation over immediate horizons. 

\subsection{Memory-based Mechanisms}\label{sec:method_memory}
To enable a learner agent ($\mathcal{A}$) to progressively infer and consolidate tacit cultural norms across sequential cross-cultural social interactions, we utilize external memory rather than weight-based methods. Acquiring implicit cultural norms requires updating hypotheses from discrete encounters without degrading models' core capabilities. External memory enables this test-time adaptation from single interactions, bypassing the catastrophic forgetting, functional interference, and computational overhead of gradient-based tuning.
Here, we evaluate two cognitively inspired memory-based mechanisms with distinct update speeds. We detail the dual-tier cultural memory architecture and its implementation in \S\ref{sec:bdi} and \S\ref{sec:memr}.

\noindent\textbf{Dual-Tier Adaptive Cultural Memory.} 
Following existing external memory architectures \cite{memory/uist/ParkOCMLB23, DBLP:journals/tois/ZhangDBMLCZDW25, suzgun-etal-2026-dynamic}, we formulate a dual-tier memory system comprising session-wise Short-Term Memory (STM) and a cross-session Long-Term Memory (LTM), described below: 

\noindent\underline{Short-Term Memory (STM).}
The STM operates as an episodic dialogue buffer recording the active turn-by-turn trajectory $\mathcal{H}_{t, k} = [u_{t, 1}, v_{t, 1}, \dots, u_{t, k-1}, v_{t, k-1}]$. The STM buffer is reset at the termination of $\mathcal{E}_t$ and serialized for post-hoc reflection and logging if required.

\noindent\underline{Long-Term Memory (LTM).}
LTM is used when normative information (i.e., cultural norms or beliefs) is available. The memory is accessed via simple retrieval and update mechanisms, and retrieved normative information is incorporated into the system prompt. The frequency of memory updates depends on the specific implementation details, which we will describe in the following sections. For clean experimental control, we exclude memory deletion and compression from the scope of the work. 

\subsubsection{BDI Memory}\label{sec:bdi}
Grounded in the Belief-Desire-Intention (BDI) framework \citep{Georgeff1999BDI} and adapted from \citet{bdi/abs-2608-22411}, this mechanism ($\text{Memory}_{\text{BDI}}$) instructs the agent $\mathcal{A}$ at each turn to infer the NPC's BDI state from the conversation history, dynamically create or update its memory records, and then generate an utterance.
$\mathcal{A}$ tracks intentions in short-term memory throughout the interaction (as they are typically interaction-specific) and optionally stores associated beliefs and desires in long-term memory at each turn.
During inference on a new interaction, the agent retrieves the top-$n$ most similar past interactions and their associated belief and desires to use as part of the system prompt.
Drawing on BDI accounts of norms \citep{criado2010normative}, we treat beliefs and desires as functional proxies for underlying norms, anchoring agent behaviour in them to foster alignment with recurring implicit cultural norms.

\subsubsection{Reflective Memory}\label{sec:memr}
A memory mechanism (Memory$_{\text{R}}$; inspired by \citealt{suzgun-etal-2026-dynamic}) that allows the model to retain and retrieve relevant cultural norms summarized from previous interactions.

At the end of each interaction session, the learner agent $\mathcal{A}$ performs inductive normative reasoning conditioned on the full transcript $\tau_t$, the situational context $\mathbf{x}_t$, and the change in social reputation score $\Delta r_t$: $\mathcal{N}_t = \mathcal{R}_{\text{norm}}(\tau_t, \mathbf{x}_t, \Delta r_t) = \{n_{t, 1}, n_{t, 2}, \dots, n_{t, M}\}$, where each $n_{t,m}$ represents an abstract, portable behavioural norm (e.g., \emph{``Always present business cards with both hands and review them attentively before storing.''}).

The extracted norms are stored in LTM alongside the interaction setting, updating the setting's associated cultural norms. Given a new interaction, the top-$n$ sets of cultural norms are retrieved to guide session behaviours as part of the system prompt. Memory$_{\text{R}}$ has a slower adaptation at the session level compared to Memory$_{\text{BDI}}$ and SR.

\section{Experimental Setup}
\label{sec:exp_setup}

\noindent\textbf{Models.}
Although the mechanisms evaluated in our paper are model-agnostic and can, in principle, be integrated into LLMs of any scale, we focus our evaluation on small open-source learner models (2–8B parameter range) because they strike a balance between real-world utility and computational efficiency \cite{smallllm, edgesurvey/corr/abs-2604-21637}, leaving the exploration of larger learner agents to future work. We examine six instruction-tuned LLMs across three model families. For the NPC agent, we select a larger model due to its superior reasoning, role-playing, and instruction-following capabilities. For efficiency, the same NPC model serves as the evaluation judge; when acting as a judge, the model retains no information regarding the NPC agent.

We use instruction-tuned models to avoid confounding poor instruction-following ability with a failure to adapt: Qwen3-4B/8B (\citealt{yang2025qwen3technicalreport}), Llama-3.2-3B/Llama-3.1-8B \citep[referred as Llama3-3B/Llama3-8B in the paper]{grattafiori2024llama3herdmodels}, and Gemma4-E2B/Gemma4-E4B \citep{gemmateam2026gemma4}. 
We select Gemma4-31B as the NPC model due to its superior instruction-following capabilities. While we also evaluated Qwen3-32B as an NPC, it underperformed relative to Gemma4-31B in our setting (detailed results in \S\ref{subsec:setup_val}); hence, we retained Gemma4-31B.
We disable thinking across all learner models where available, to maintain a fair comparison (e.g., Llama3 models are non-reasoning models) and to isolate the effects of reasoning prompts, memory, and the reflection mechanism. Our main experimental results are averaged over three runs with statistical testing. Models are accessed via Hugging Face Transformers \citep{wolf-etal-2020-transformers}.

\noindent\textbf{Methods and Mechanisms.} We evaluate all the methods and mechanisms outlined in \S\ref{sec:methods}, as well as the default base models.
For efficiency, retrievals for both $\text{Memory}_{\text{BDI}}$ and $\text{Memory}_{\text{R}}$ are precomputed.
The session contexts are embedded into a dense semantic space via Gemini Embedding-2 \cite{gemini_embedding2}. For every interaction, both methods retrieve norms from the top-$n$ most similar sessions via cosine similarity above a set threshold.

\noindent\textbf{Hyperparameters.} 
Main experimental results are averaged across three runs and supported by statistical tests. For standard repeated interactions, we evaluate 50 three-episode scenarios (sessions) per culture, withholding the remaining scenarios for generalization experiments as unseen scenarios (\S\ref{sec:gen}).
The order of the scenarios is randomized at the beginning of each episode. For experiments with a memory, results are further averaged over episodes 2 and 3 (due to longitudinal learning), and with top-2 retrieval (with the first one always being the same scenario). The social reputation ($[-1, 1]$) scores are tracked and averaged over the scenarios in each episode.

We set the maximum interaction turns to 4, since prior work has shown that LLM persona consistency degrades over extended role-play \cite{li2024measuring, DBLP:conf/iclr/LabanHZN26}. The decoding temperature is set to 0.7 for learner models ($\mathcal{A}$), 0.4 for the NPC model during interactions (maintaining low output diversity without pure determinism to ensure consistency across repeated interactions), and 0 when the NPC model functions as a judge.

\subsection{Experimental Setup Validation}
\label{subsec:setup_val}
To validate the reliability of our simulated environment prior to full-scale evaluation, we compared two candidate NPC/Judge models (Gemma4-31B and Qwen3-32B) on their judging quality and instruction adherence. We tested both candidates across two environments by pairing them with a representative learner agent from each model family (Qwen3-8B, Llama3-8B, and Gemma4-E4B) using methods outlined in \S\ref{sec:exp_setup}.

We first sampled 60 interactions per environment (120 in total), stratified by learner model. Overall, 62.5\% were judged as having positive social reputation by models. We evaluated the judgment accuracy of social norm adherence (i.e., social reputation score). Sampled scenarios were independently verified by two human evaluators (marked as wrong judgment if either evaluator considered it wrong). In the Gemma NPC environment, human evaluators accepted 91.67\% of social-norm judgments across 60 sampled interactions. For the Qwen NPC environment, these rates were 73.3\%. Overall, human verification results indicate that the Gemma4-31B NPC yield more reliable outcomes than the Qwen NPC, justifying its selection for subsequent experiments. 

Additionally, we evaluated NPC norm leakage using GPT-6-Luna \cite{openai2026gpt6luna} across 60 sampled positive-reputation interactions with Gemma NPCs, which revealed a low dialogue leakage in 8.3\% of cases.

\section{Results and Analysis}

\subsection{Main Results}

\begin{table*}[t]
\centering
\small
\resizebox{0.92\textwidth}{!}{%
\begin{tabular}{l cccccl}
\toprule
\textbf{Model} & \textbf{CN} & \textbf{FR} & \textbf{JP} & \textbf{NZ} & \textbf{SA} & \textbf{Avg.}\\
\midrule
Gemma4-E4B                             & \goalc{0.01} $\pm$ 0.04 & \goalc{0.22} $\pm$ 0.13 & \goalc{-0.05} $\pm$ 0.01 & \goalc{0.33} $\pm$ 0.08 & \goalc{0.15} $\pm$ 0.02 & \srepc{0.13} \\
Gemma4-E4B SR                          & \goalc{0.00} $\pm$ 0.06 & \goalc{0.06} $\pm$ 0.03 & \goalc{-0.06} $\pm$ 0.04 & \goalc{0.42} $\pm$ 0.03 & \goalc{0.25} $\pm$ 0.03 & \srepc{0.13} \\
Gemma4-E4B $_{\text{Memory}_{\text{BDI}}}$ & \goalc{0.04} $\pm$ 0.05 & \goalc{0.36} $\pm$ 0.02 & \goalc{0.10} $\pm$ 0.05 & \goalc{0.53} $\pm$ 0.05 & \goalc{0.23} $\pm$ 0.03 & \srepc{0.25}* \\
Gemma4-E4B $_{\text{Memory}_{\text{R}}}$ & \goalc{0.23} $\pm$ 0.04 & \goalc{0.38} $\pm$ 0.02 & \goalc{0.23} $\pm$ 0.06 & \goalc{0.51} $\pm$ 0.12 & \goalc{0.35} $\pm$ 0.04 & \srepc{0.34}* \\
\midrule

Qwen3-8B                               & \goalc{-0.47} $\pm$ 0.03 & \goalc{-0.24} $\pm$ 0.02 & \goalc{-0.27} $\pm$ 0.06 & \goalc{0.21} $\pm$ 0.05 & \goalc{-0.11} $\pm$ 0.10 & \srepc{-0.17} \\
Qwen3-8B SR                            & \goalc{-0.23} $\pm$ 0.02 & \goalc{-0.11} $\pm$ 0.07 & \goalc{-0.08} $\pm$ 0.00 & \goalc{0.27} $\pm$ 0.06 & \goalc{-0.06} $\pm$ 0.03 & \srepc{-0.04}* \\
Qwen3-8B $_{\text{Memory}_{\text{BDI}}}$ & \goalc{-0.17} $\pm$ 0.05 & \goalc{-0.03} $\pm$ 0.03 & \goalc{-0.18} $\pm$ 0.04 & \goalc{0.30} $\pm$ 0.04 & \goalc{0.07} $\pm$ 0.05 & \srepc{-0.00}* \\
Qwen3-8B $_{\text{Memory}_{\text{R}}}$   & \goalc{-0.10} $\pm$ 0.03 & \goalc{0.03} $\pm$ 0.02 & \goalc{-0.12} $\pm$ 0.01 & \goalc{0.24} $\pm$ 0.03 & \goalc{0.01} $\pm$ 0.13 & \srepc{0.01}* \\
\midrule
Llama3-8B                              & \goalc{-0.29} $\pm$ 0.10 & \goalc{-0.13} $\pm$ 0.08 & \goalc{-0.27} $\pm$ 0.09 & \goalc{0.18} $\pm$ 0.02 & \goalc{-0.15} $\pm$ 0.15 & \srepc{-0.13} \\
Llama3-8B SR                           & \goalc{-0.31} $\pm$ 0.03 & \goalc{-0.05} $\pm$ 0.02 & \goalc{-0.26} $\pm$ 0.10 & \goalc{0.14} $\pm$ 0.04 & \goalc{-0.01} $\pm$ 0.08 & \srepc{-0.10} \\
Llama3-8B $_{\text{Memory}_{\text{BDI}}}$ & \goalc{-0.42} $\pm$ 0.11 & \goalc{-0.23} $\pm$ 0.06 & \goalc{-0.35} $\pm$ 0.06 & \goalc{-0.03} $\pm$ 0.02 & \goalc{-0.10} $\pm$ 0.02 & \srepc{-0.23} \\
Llama3-8B $_{\text{Memory}_{\text{R}}}$  & \goalc{-0.21} $\pm$ 0.10 & \goalc{0.07} $\pm$ 0.08 & \goalc{-0.18} $\pm$ 0.05 & \goalc{0.17} $\pm$ 0.01 & \goalc{0.01} $\pm$ 0.03 & \srepc{-0.03}* \\

\midrule
\midrule
Gemma4-E2B                             & \goalc{-0.25} $\pm$ 0.03 & \goalc{0.01} $\pm$ 0.16 & \goalc{-0.28} $\pm$ 0.10 & \goalc{0.27} $\pm$ 0.05 & \goalc{-0.04} $\pm$ 0.00 & \srepc{-0.06} \\
Gemma4-E2B SR                          & \goalc{-0.35} $\pm$ 0.01 & \goalc{-0.11} $\pm$ 0.03 & \goalc{-0.29} $\pm$ 0.05 & \goalc{0.15} $\pm$ 0.11 & \goalc{-0.05} $\pm$ 0.11 & \srepc{-0.13} \\
Gemma4-E2B $_{\text{Memory}_{\text{BDI}}}$ & \goalc{-0.27} $\pm$ 0.01 & \goalc{-0.09} $\pm$ 0.10 & \goalc{-0.28} $\pm$ 0.07 & \goalc{0.41} $\pm$ 0.09 & \goalc{0.14} $\pm$ 0.06 & \srepc{-0.02} \\
Gemma4-E2B $_{\text{Memory}_{\text{R}}}$ & \goalc{0.02} $\pm$ 0.08 & \goalc{0.15} $\pm$ 0.13 & \goalc{0.02} $\pm$ 0.00 & \goalc{0.38} $\pm$ 0.16 & \goalc{0.24} $\pm$ 0.04 & \srepc{0.16}* \\

\midrule
Qwen3-4B                               & \goalc{-0.26} $\pm$ 0.03 & \goalc{-0.19} $\pm$ 0.07 & \goalc{-0.21} $\pm$ 0.09 & \goalc{0.40} $\pm$ 0.03 & \goalc{-0.11} $\pm$ 0.11 & \srepc{-0.07} \\
Qwen3-4B SR                            & \goalc{-0.28} $\pm$ 0.05 & \goalc{-0.17} $\pm$ 0.02 & \goalc{-0.18} $\pm$ 0.03 & \goalc{0.27} $\pm$ 0.09 & \goalc{0.19} $\pm$ 0.05 & \srepc{-0.03} \\
Qwen3-4B $_{\text{Memory}_{\text{BDI}}}$ & \goalc{-0.23} $\pm$ 0.06 & \goalc{-0.01} $\pm$ 0.05 & \goalc{-0.19} $\pm$ 0.03 & \goalc{0.37} $\pm$ 0.00 & \goalc{0.06} $\pm$ 0.04 & \srepc{-0.00}* \\
Qwen3-4B $_{\text{Memory}_{\text{R}}}$   & \goalc{-0.18} $\pm$ 0.04 & \goalc{0.07} $\pm$ 0.04 & \goalc{-0.07} $\pm$ 0.08 & \goalc{0.37} $\pm$ 0.03 & \goalc{0.08} $\pm$ 0.09 & \srepc{0.05}* \\

\midrule
Llama3-3B                              & \goalc{-0.27} $\pm$ 0.04 & \goalc{-0.15} $\pm$ 0.03 & \goalc{-0.38} $\pm$ 0.00 & \goalc{0.17} $\pm$ 0.10 & \goalc{-0.25} $\pm$ 0.03 & \srepc{-0.18} \\
Llama3-3B SR                           & \goalc{-0.29} $\pm$ 0.07 & \goalc{-0.11} $\pm$ 0.03 & \goalc{-0.28} $\pm$ 0.11 & \goalc{0.13} $\pm$ 0.02 & \goalc{-0.07} $\pm$ 0.04 & \srepc{-0.13} \\
Llama3-3B $_{\text{Memory}_{\text{BDI}}}$ & \goalc{-0.37} $\pm$ 0.06 & \goalc{-0.30} $\pm$ 0.05 & \goalc{-0.38} $\pm$ 0.04 & \goalc{-0.03} $\pm$ 0.03 & \goalc{-0.09} $\pm$ 0.07 & \srepc{-0.23} \\
Llama3-3B $_{\text{Memory}_{\text{R}}}$  & \goalc{-0.16} $\pm$ 0.11 & \goalc{-0.01} $\pm$ 0.20 & \goalc{-0.27} $\pm$ 0.16 & \goalc{0.26} $\pm$ 0.11 & \goalc{-0.13} $\pm$ 0.13 & \srepc{-0.06}* \\
\bottomrule
\end{tabular}}
\caption{Model performance across cultures: China (CN), France (FR), Japan (JP), New Zealand (NZ), and Saudi Arabia (SA). Social Reputation $\in [-1.0, 1.0]$: episode-averaged score measuring adherence to cultural norms ($-1.0$ indicates any violations or cultural faux pas). $^*$: a statistically significant improvement over the default ($p < 0.05$).}
\label{tab:main}
\end{table*}

Table \ref{tab:main} shows models' social reputations during the interactions.  Across the evaluated mechanisms, situational reasoning (SR) shows only inconsistent gains in social reputation. 
Furthermore, Memory${_\text{BDI}}$ often improves performance but fails to generalize across model families (e.g., Llama 3 models). In contrast, Memory${_\text{R}}$ is the most effective mechanism, consistently achieving the highest social reputation scores across all models over the default baselines (with a considerable gain of +0.21 on Gemma4-E4B, where social reputation is measured on a scale of $[-1,1]$). Among all evaluated models, Gemma4-E4B demonstrates the best overall performance in retaining a positive social reputation, whereas Llama models maintain a negative social reputation on average with almost all mechanisms. In general, larger models within the same family outperform their smaller counterparts (also confirmed by a Qwen3-32B experiment in Table \ref{ref:large}, Appendix). This is likely because larger models possess stronger baseline knowledge and capabilities, and are better at norms utilization (as also supported by our analysis in Figure \ref{fig:norm_semantics}).

Furthermore, across all models and methods, performance was highest for New Zealand culture, whereas Chinese and Japanese cultures consistently received the lowest social reputation scores.
Such consistent patterns suggest that the observed differences could reflect biases embedded in the models' training data or underlying representations of different cultures, which may affect social tasks such as collaboration and negotiation, warranting investigation in future work.

Individual statistical analyses comparing SR, Memory${_\text{BDI}}$, and Memory${_\text{R}}$ with the default reveal that Memory$_{\text{R}}$ has a statistically significant advantage over the default for all models ($p=0.0020$ for Llama3-3B, and $p=0.0000$ for all other models). Memory${_\text{BDI}}$ also shows a statistically significant advantage over the default, except for the Llama3 models on social reputation scores.

\begin{figure*}[h!]
    \centering

    \begin{subfigure}{0.95\linewidth}
        \centering
        \includegraphics[width=\linewidth]{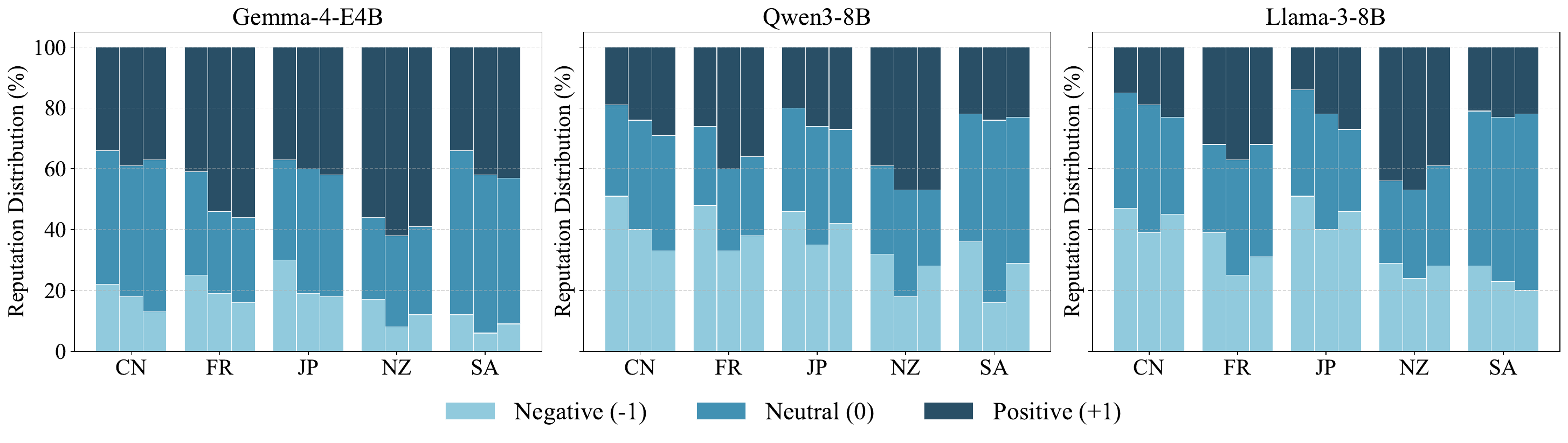}
        \label{fig:main_large}
    \end{subfigure}


    \begin{subfigure}{0.95\linewidth}
        \centering
        \includegraphics[width=\linewidth]{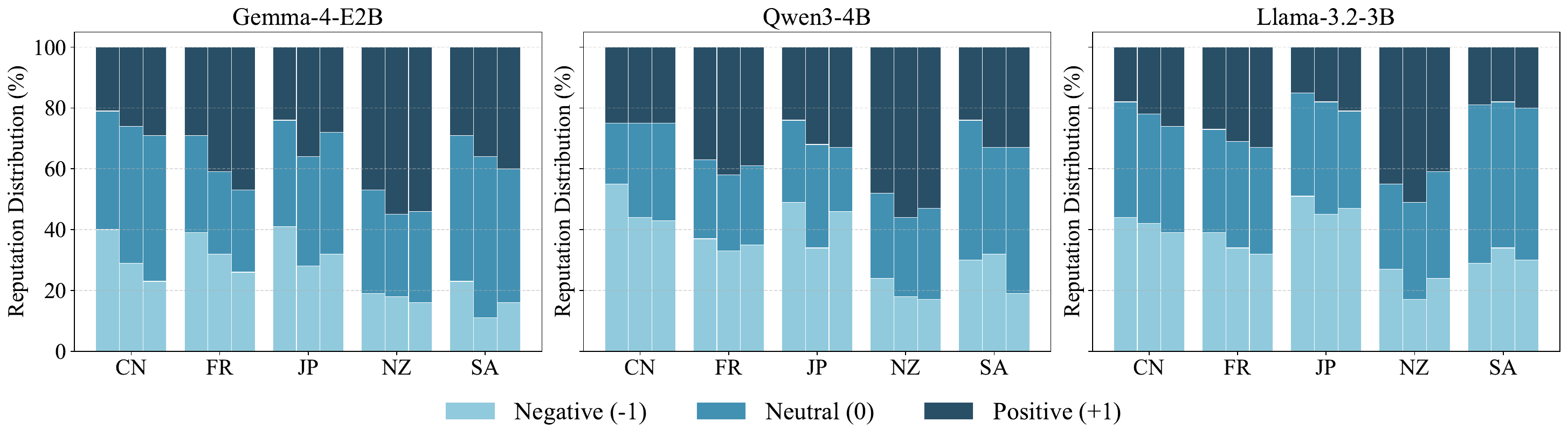}
        \label{fig:main_small}
    \end{subfigure}

    \caption{Distribution of social reputation judgments under $\text{Memory}_{\text{R}}$ across models and cultures for English-language interactions. For each cultural group, the three bars (left to right) represent sequential repeated interaction episodes. Stacked segments denote the proportion of scenarios judged as Positive (+1; norm-aligning), Neutral (0; partially adhering), or Negative (-1; norm-violating). China (CN), France (FR), Japan (JP), New Zealand (NZ), and Saudi Arabia (SA).}
    \label{fig:longitudinal}
\end{figure*}

\subsection{Longitudinal Improvement}
Between the two memory mechanisms, $\text{Memory}_{\text{R}}$ demonstrates more consistent longitudinal improvement than $\text{Memory}_{\text{BDI}}$.

Figure~\ref{fig:longitudinal} shows the distribution of social reputation scores across three episodes under $\text{Memory}_{\text{R}}$. 
Across all models, positive scores generally increase by Episode 2. However, a third episode does not consistently yield further improvements, though performance remains above baseline. 
This pattern likely reflects the models' sensitivity to prompts, leading to norm revisions that are less conducive to maintaining a positive reputation.

A statistical analysis also confirms that $\text{Memory}_{\text{R}}$ yields statistically significant longitudinal gains (Table~\ref{app:tab:longitudinal}). Specifically, larger models exhibit significant within-encounter improvements ($\text{Post} - \text{Ep.1}$, where \text{Post} is the averaged result between episode 2 and 3.) under $\text{Memory}_{\text{R}}$ for Gemma4-E4B ($\Delta = +0.15$, $p < 0.0001$), Qwen3-8B ($\Delta = +0.18$, $p < 0.0001$), and Llama3-8B ($\Delta = +0.11$, $p = 0.0007$). Smaller models follow a similar pattern, showing significant gains for Gemma4-E2B ($\Delta = +0.18$, $p < 0.0001$), Qwen3-4B ($\Delta = +0.12$, $p = 0.0002$), and Llama3-3B ($\Delta = +0.07$, $p = 0.0094$). When evaluated using a linear mixed model (LMM), Gemma4-E4B ($\beta = +0.10$, $p = 0.0016$), Qwen3-8B ($\beta = +0.08$, $p = 0.0161$), Llama3-8B ($\beta = +0.07$, $p = 0.0220$), and Llama3-3B ($\beta = +0.06$, $p = 0.0209$) also demonstrate significant positive longitudinal trajectories under $\text{Memory}_{\text{R}}$.

In contrast, despite higher episode 1 performance across all Gemma and Qwen models (Table~\ref{tab:main}), $\text{Memory}_{\text{BDI}}$ yields significant longitudinal learning only in Qwen3-4B ($p = 0.0099$).
This trade-off between initial performance and longitudinal gain is expected, as $\text{Memory}_{\text{BDI}}$ adapts at each turn while $\text{Memory}_{\text{R}}$ consolidates norms post-session. Although both receive social reputation scores, $\text{Memory}_{\text{R}}$ uses it as an implicit signal to revise norms, whereas $\text{Memory}_{\text{BDI}}$ likely relies more on turn-level interaction.

These results show that while memory facilitates cultural norm assimilation in general, frequent updates ($\text{Memory}_{\text{BDI}}$) of normative information offer no advantage, likely due to amplifying local noise.

\begin{table}[t]
\centering
\small
\resizebox{0.95\columnwidth}{!}{%
\begin{tabular}{l ccccc}
\toprule
\textbf{Model} & \textbf{Ep.1} & \textbf{Post} & \textbf{$p_{\text{within}}$} & \textbf{$\beta$} & \textbf{$p_{\beta}$} \\
\midrule
\multicolumn{6}{l}{\textit{Gemma4-E4B}} \\
\quad $\text{Memory}_{\text{BDI}}$ & $+0.22$ & $+0.25$ & $0.1380$ & $+0.05$ & $0.1260$ \\
\quad $\text{Memory}_{\text{R}}$ & $+0.19$ & $+0.34$ & $<0.0001^*$ & $+0.10$ & $0.0016^*$ \\
\addlinespace[3pt]
\multicolumn{6}{l}{\textit{Qwen3-8B}} \\
\quad $\text{Memory}_{\text{BDI}}$ & $-0.01$ & $-0.00$ & $0.3838$ & $+0.01$ & $0.7241$ \\
\quad $\text{Memory}_{\text{R}}$ & $-0.17$ & $+0.01$ & $<0.0001^*$ & $+0.08$ & $0.0161^*$ \\
\addlinespace[3pt]
\multicolumn{6}{l}{\textit{Llama3-8B}} \\
\quad $\text{Memory}_{\text{BDI}}$ & $-0.24$ & $-0.23$ & $0.2580$ & $+0.04$ & $0.1804$ \\
\quad $\text{Memory}_{\text{R}}$ & $-0.14$ & $-0.03$ & $0.0007^*$ & $+0.07$ & $0.0220^*$ \\
\addlinespace[3pt]
\multicolumn{6}{l}{\textit{Gemma4-E2B}} \\
\quad $\text{Memory}_{\text{BDI}}$ & $-0.01$ & $-0.01$ & $0.6119$ & $-0.04$ & $0.2139$ \\
\quad $\text{Memory}_{\text{R}}$ & $-0.02$ & $+0.16$ & $<0.0001^*$ & $+0.06$ & $0.0688$ \\
\addlinespace[3pt]
\multicolumn{6}{l}{\textit{Qwen3-4B}} \\
\quad $\text{Memory}_{\text{BDI}}$ & $-0.06$ & $0.00$ & $0.0099^*$ & $+0.02$ & $0.4105$ \\
\quad $\text{Memory}_{\text{R}}$ & $-0.07$ & $+0.05$ & $0.0002^*$ & $+0.05$ & $0.0853$ \\
\addlinespace[3pt]
\multicolumn{6}{l}{\textit{Llama3-3B}} \\
\quad $\text{Memory}_{\text{BDI}}$ & $-0.24$ & $-0.23$ & $0.3455$ & $+0.03$ & $0.2675$ \\
\quad $\text{Memory}_{\text{R}}$ & $-0.13$ & $-0.06$ & $0.0094^*$ & $+0.06$ & $0.0209^*$ \\
\bottomrule
\end{tabular}}
\caption{Longitudinal social reputation performance across interaction episodes. Evaluated across $N=250$ paired scenario encounters across all cultures. \textbf{Ep.1}: Episode~1. \textbf{Post}: average performance of Ep.2 and Ep.3. $\boldsymbol{\beta}$: the fixed-effect longitudinal trajectory slope from a linear mixed model (LMM), with significance $p_{\beta}$. $^*$ indicates statistically significant ($p < 0.05$).}
\label{app:tab:longitudinal}
\end{table}

\subsection{Inferred Norms versus Ground Truth}
While models exhibit statistically significant improvement, how closely the recovered cultural norms match the ground truth remains unknown. In this section, we compare norms inferred by $\text{Memory}_{\text{R}}$ to ground-truth norms across each culture and setting.

\begin{figure*}[]
    \centering
    \includegraphics[width=0.95\linewidth]{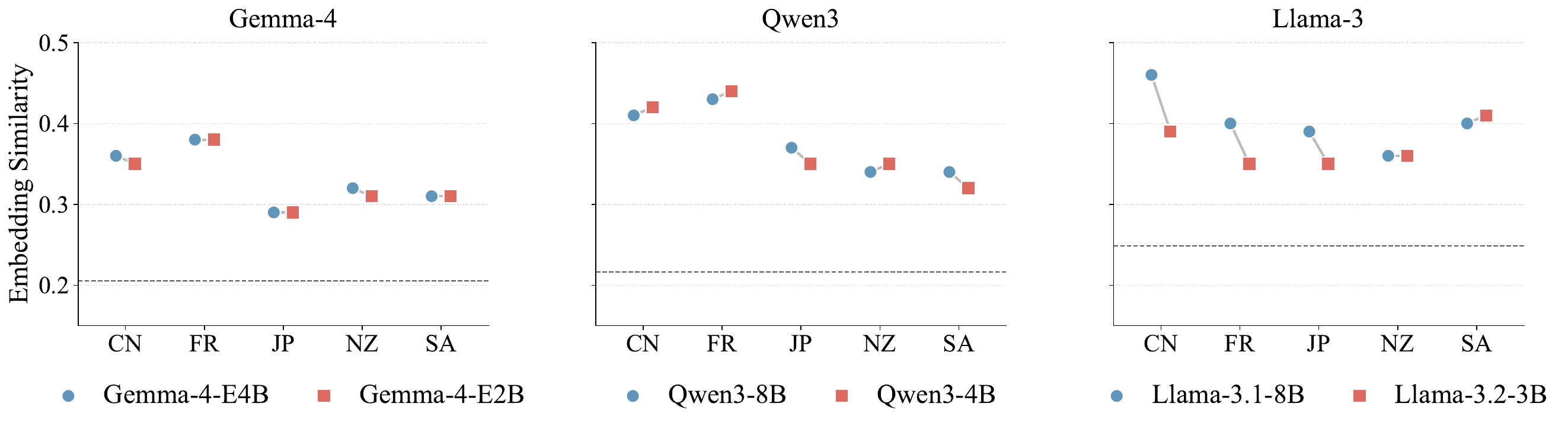}
    \caption{Embedding similarity between inferred norms and ground truth across models and cultures under $\text{Memory}_{\text{R}}$. 
    The dashed line indicates the background embedding similarity for randomly paired norms for that model family.}
    \label{fig:norm_semantics}
\end{figure*}

Figure \ref{fig:norm_semantics} shows the embedding similarities between the inferred and summarized norms and the ground-truth norms. Overall, all models except Llama-3.2-3B exhibit higher similarity to the French and Chinese ground-truth norms and lower similarity to the Japanese and New Zealand norms. All norms show non-trivial embedding similarity with respect to the ground truth, substantially exceeding the background baseline from randomly paired norms within culture for that model family.
Except for Llama3 models, negligible within-family size differences suggest that the performance disparities reflect models' ability to utilize norms rather than to infer them. This inference-execution gap highlights the need to move beyond recall-based evaluation.

Table~\ref{tab:exp} (appendix) provides qualitative examples of Gemma4-E4B's norm outputs.

\subsection{Generalization to Different Scenarios}\label{sec:gen}

\begin{figure}[h!]
    \centering
    \includegraphics[width=0.98\columnwidth]{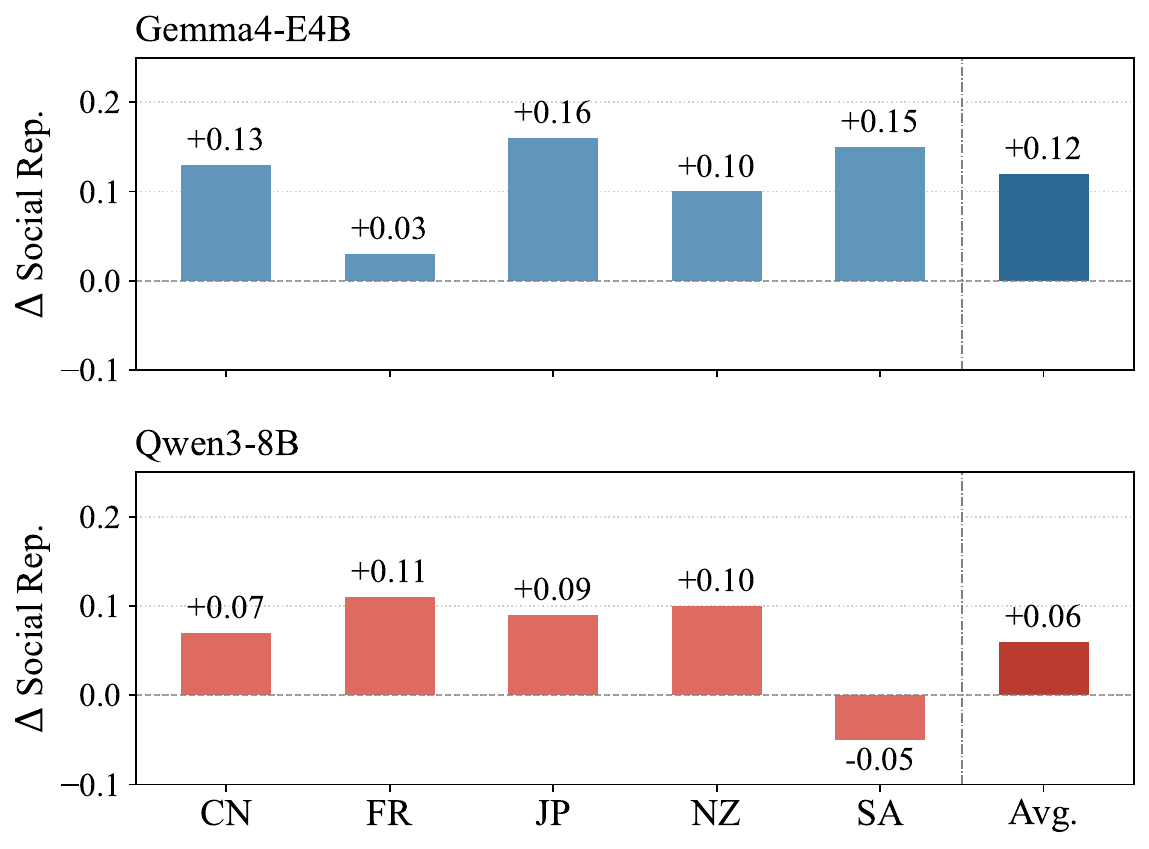}
    \caption{Change in social reputation scores on unseen scenarios using norms inferred from repeated interactions ($\text{Memory}_{\text{R}}$) relative to the default.}
    \label{fig:generalization}
\end{figure}

Using the norms inferred under $\text{Memory}_{\text{R}}$ for each culture of Gemma4-E4B and Qwen3-8B (two most improved models), we evaluate the models on held-out scenarios to test their generalized behaviour (without repeated interactions). The gains in social reputation compared with using the corresponding default models are shown in Figure \ref{fig:generalization}.

Overall, compared to the default, $\text{Memory}_{\text{R}}$ improves social reputation by 0.12 points and 0.06 points for Gemma4-E4B and Qwen3-8B, respectively. Notably, both models consistently improved performance across cultures, except for Saudi Arabia in Qwen3-8B. Both models demonstrated clear generalization to settings in Japan, China and New Zealand. However, while Gemma4-E4B showed improvements for Saudi Arabia, this was not observed in Qwen3-8B, indicating that generalization still depends on the base model and culture.

\subsection{Multilingual Experiment}

In the multilingual experiment, we instructed the model to simulate social interactions in the primary language associated with the evaluated cultures (excluding New Zealand) with Gemma4-E4B and Qwen3-8B. The social setting, behavioural instructions, judge and norm-summarization prompts were all maintained in English due to the models' superior English-language capabilities.

Table~\ref{tab:xling} presents the results for both the baseline model and $\text{Memory}_{\text{R}}$. Consistent with previous English-language results, $\text{Memory}_{\text{R}}$ leads to higher averaged social reputation scores across cultures. Across languages, Gemma4-E4B achieves higher performance in Arabic and French, while Qwen3-8B performs better specifically in French.

\begin{table}[]
\centering
\small
\resizebox{0.98\columnwidth}{!}{
\begin{tabular}{lcccc c}
    \toprule
\textbf{Model} & \textbf{CN} & \textbf{FR} & \textbf{JP} & \textbf{SA} & \textbf{Avg.}\\
\midrule
Gemma4-E4B & \goalc{-0.06} & \goalc{0.21} & \goalc{0.07} & \goalc{0.25} & \srepc{0.12} \\
Gemma4-E4B $_{\text{Memory}_{\text{R}}}$ & \goalc{0.23} & \goalc{0.34} & \goalc{0.20} & \goalc{0.54} & \srepc{0.33} \\
Qwen3-8B & \goalc{-0.21} & \goalc{-0.05} & \goalc{-0.14} & \goalc{-0.18} & \srepc{-0.15} \\
Qwen3-8B $_{\text{Memory}_{\text{R}}}$   & \goalc{-0.06} & \goalc{-0.04} & \goalc{-0.07} & \goalc{-0.07} & \srepc{-0.06} \\
\bottomrule
\end{tabular}}
\caption{Averaged multilingual performance (social reputation score, $\text{Memory}_{\text{R}}$) across four cultures: China (CN), France (FR), Japan (JP), and Saudi Arabia (SA).}
\label{tab:xling}
\end{table}

Compared to baseline interactions, $\text{Memory}_{\text{R}}$ demonstrated statistically significant social reputation across episodes with both Gemma4-E4B ($p = 0.0001$) and Qwen3-8B ($p=0.0311 $) compared to the default. While longitudinal learning is observed, LMM per-episode slopes were not significant (Gemma: $\beta = +0.07$ pts/ep, $p = 0.0591$; Qwen: $\beta =+0.04$ pts/ep, $p=0.3109$), such that evidence for a consistent longitudinal trend is marginal for Gemma and absent for Qwen.

Figure \ref{fig:xling} (appendix) visualizes the longitudinal results of multilingual experiments.

\subsection{Generalization to Synthetic Norms}\label{sec:synth-norm}

To test whether $\text{Memory}_{\text{R}}$ generalizes to counter-intuitive, synthetic cultural norms, we conduct a preliminary experiment containing 15 scenarios grounded in hand-revised human norms (manually rewritten into a counterfactual form; see Appendix~\ref{app:synth-norm} for creation details). For example, the original human norm of ``Pay respect to seniors in all situations.'' was transformed into ``Pay respect to \emph{children} in all situations.'' To prevent the model from relying on culture-specific priors, all NPCs use a single generic identifier (\emph{Agent 12345}) and their cultural identity was set to \emph{Unknown}.

Overall, Gemma4-E4B under Memory$_R$ consistently outperforms the default setting, achieving a social reputation score of -0.033 (default: -0.422).
We also observe a positive longitudinal learning slope across episodes, although the trend is not statistically significant ($\beta = +0.20$ pts/ep, $p = 0.1405$). This may stem from the limited number of evaluation scenarios, as well as the inherent difficulty of adapting to counterfactual norms. Nonetheless, these findings suggest that memory augmentation benefits agent behaviour in social contexts, though further work is needed to enhance progressive learning in unfamiliar situations.

\subsection{Longer Interaction Horizons}

\begin{figure}
    \centering
    \includegraphics[width=0.95\linewidth]{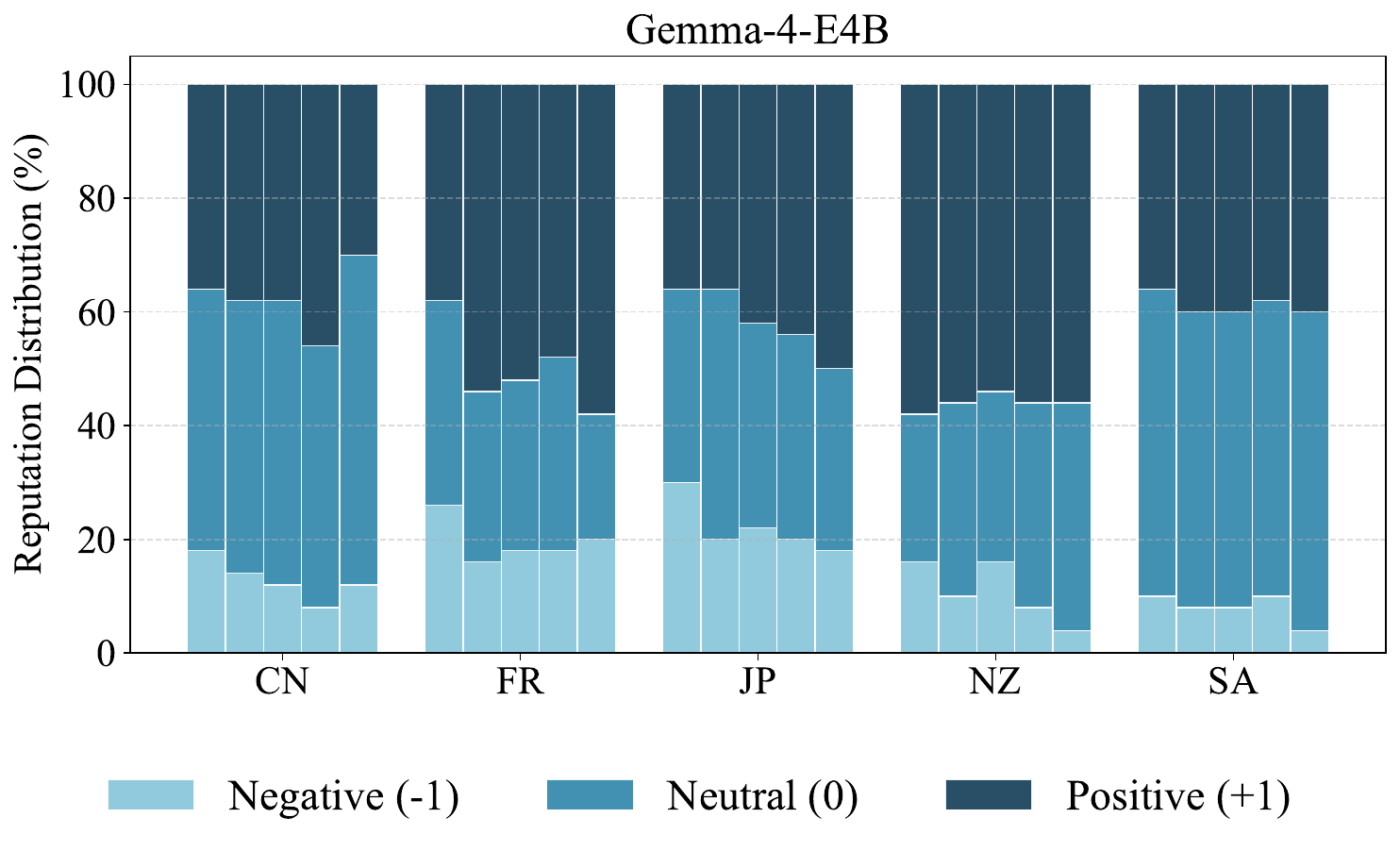}
    \caption{Distribution of social reputation judgments under $\text{Memory}_{\text{R}}$ for 5 episodes. For each culture, the bars (left to right) represent sequential repeated interaction episodes. Stacked segments denote the proportion of scenarios judged as Positive (+1; norm-aligning), Neutral (0; partially adhering), or Negative (-1; norm-violating).}
    \label{fig:long}
\end{figure}

We further evaluated Gemma4-E4B with $\text{Memory}_{\text{R}}$ across five episodes in English to assess whether adaptive memory maintains its benefits over longer horizons.
The model demonstrates statistically significant longitudinal improvement in social reputation across repeated sequential encounters (Figure \ref{fig:long}), improving from an initial mean of $+0.21\text{ pts}$ at episode 1 to $+0.35\text{ pts}$ by episode 5. This yields a statistically significant longitudinal gain ($\Delta_{\text{ep.5}-\text{ep.1}} = +0.14\text{ pts}$, $95\%\text{ CI } [+0.04, +0.24]$, $p = 0.0026$), as well as a sustained post-adaptation horizon gain across episode 2--5 ($\Delta_{\text{post}} = +0.12\text{ pts}$, $95\%\text{ CI } [+0.04, +0.20]$, $p = 0.0028$). Evaluated with a linear mixed-effects trend model, the method exhibits a significant positive learning trajectory ($\beta = +0.031\text{ pts/ep}$, $p = 0.0140$). 

These results point to memory mechanisms as a promising way to support the continuous norm assimilation over extended horizons.

\section{Conclusion}
Grounded in human culture and social learning, this work investigated LLMs' capacity to assimilate implicit cultural norms and sustain culturally congruent behaviours across ongoing social interactions using lightweight memory mechanisms. Using our five-culture social interactive sandbox, we evaluated longitudinal cultural adaptation across six open-source models across languages.

Our results provide three central takeaways. First, we show that augmenting models with external memory enhances adaptive plasticity and generalization, enabling continuous norm assimilation across languages. Second, we find that session-level memory consolidation is more effective than turn-level adaptation, suggesting that how interaction history is consolidated matters for sustained cultural adaptation. 
Third, we reveal an inference-execution gap in smaller models, indicating that norm inference and recognition alone do not guarantee behavioural alignment.

Overall, these results broaden the scope of cultural adaptation beyond knowledge recall and demonstrate the feasibility of interaction-driven cultural adaptation. By showing that cultural norms can be inferred through social interaction driven by environmental incentives, our problem formulation and findings lay the groundwork for future human- and cognitively inspired approaches to continuous cultural adaptation.

\section*{Limitations}

In our work, the scenario data was generated using Gemini models, which could introduce a confound where Gemma 4 models achieve higher absolute performance as learner agents due to familiarity with the writing style. Nevertheless, because we evaluated multiple learner model families and found consistent learning patterns, this stylistic familiarity does not compromise our primary objective and conclusions.

In our work, cultural boundaries are drawn at the country level. Because our scenarios are derived from prior human norm datasets, inheriting such scope is necessary, and facilitating easy comparisons. Our analysis remains sufficiently general to support finer-grained divisions, such as regional or subcultural groups, although this is beyond the scope of the current work. 

Furthermore, social scenarios can be formulated through mechanisms beyond using using social goals as the motivation. While our work establishes a foundational starting point, future research could extend this paradigm to open-ended, non-goal-directed interactions and diverse social environments.

\section*{Acknowledgement}

This work was supported by the UK Research and Innovation (UKRI) Frontier Research Grant EP/Y031350/1 EQUATE and the Isambard AI National AI Research Resource (AIRR). 

\bibliography{anthology}
\bibliographystyle{acl_natbib}

\clearpage
\appendix
\section{Prompts}\label{app:prompts}

Prompts are provided in this section. 
\textcolor{blue}{Blue text} in the prompts indicates optional text that depends on the experimental setup (e.g., whether a social reputation score is included or whether the experiment is multilingual). \textcolor{tableblue}{Light blue text} is used for Memory$_R$, while \textcolor{orange}{salmon coloured text} is used for Memory$_\text{BDI}$.

\begin{figure*}[t]
\centering
\small
\begin{promptbox}[title={Cultural Norm Leakage Evaluator Prompt}, label={prompt:leakage}]
You are an expert evaluator assessing whether ground truth cultural norms were leaked, directly revealed, or artificially resolved during an interactive cultural simulation between Agent A (the main agent) and an NPC.

\medskip\noindent
Cultural Group: \texttt{\{culture\}}\\
Setting: \texttt{\{setting\}}\\
Goal: \texttt{\{goal\}}

\medskip\noindent
\textbf{Ground Truth Norms:}\\
\texttt{\{gt\_norms\_formatted\}}

\medskip\noindent
\textbf{Interaction Transcript:}\\
\texttt{\{transcript\_formatted\}}

\medskip\noindent
\textbf{Evaluation Task:}\\
Analyze whether any of the Ground Truth Norms were leaked, exposed, or artificially resolved during the interaction.

\medskip\noindent
Leakage occurs if any of the following are true:
\begin{enumerate}[leftmargin=*,itemsep=1pt,topsep=2pt]
  \item \textbf{NPC Dialogue Leak} (\texttt{"npc\_dialogue\_leak"}): The NPC directly states, quotes, instructs, or explicitly gives away the exact cultural rule, hidden requirement, or expected behavioral test in dialogue.
  \item \textbf{NPC Stage Direction / Narrative Attribution} (\texttt{"npc\_narration\_leak"}): The NPC's turn narration or action cues explicitly resolve a cultural requirement by narrating Agent A's behavior or physical actions (e.g., describing Agent A eating food, handling utensils, bowing, or placing items) that Agent A never explicitly generated or stated in their own turns.
\end{enumerate}

\medskip\noindent
\textbf{Important Instructions:}
\begin{itemize}[leftmargin=*,itemsep=1pt,topsep=2pt]
  \item Carefully check if the NPC's narration hallucinated/imputed an etiquette action for Agent A (e.g., finishing fruit, clapping) that satisfies a ground truth norm. If Agent A never wrote this action themselves, classify this as a leak (\texttt{npc\_narration\_leak}).
  \item If no leakage occurs for a norm, set \texttt{"is\_leaked"} to false and \texttt{"leakage\_type"} to \texttt{"none"}.
\end{itemize}

\medskip\noindent
For EACH Ground Truth Norm, provide:
\begin{itemize}[leftmargin=*,itemsep=1pt,topsep=2pt]
  \item \texttt{"norm\_idx"}: integer (1-based index)
  \item \texttt{"is\_leaked"}: boolean (true if leaked, false if not leaked)
  \item \texttt{"leakage\_type"}: \texttt{"npc\_dialogue\_leak"} $\vert$ \texttt{"npc\_narration\_leak"} $\vert$ \texttt{"none"}
  \item \texttt{"reason"}: concise 1 sentence explanation citing exact turn numbers and quoted text
\end{itemize}

\medskip\noindent
Return ONLY valid JSON matching this schema:

{
  "evaluations": [
    {
      "norm\_idx": 1,
      "is\_leaked": <boolean>,
      "leakage\_type": "none",
      "reason": "..."
    },
    ...
  ]
}
\end{promptbox}
\end{figure*}

\begin{figure}
\centering
\small
\resizebox{0.9\columnwidth}{!}{
\begin{promptbox}[title={Judge Prompts}, label={prompt:judge}]
\textbf{System:}\\
You are an impartial cultural behavior evaluator. \\
Your task is to assess whether Agent A's behavior aligns with \texttt{\{cultural\_group\}} social expectations.\\

\textbf{User:}\\
Setting: \texttt{\{setting\}}\\
Cultural Norm Checklist to Verify:\\
\texttt{\{checklist\}}\\
Conversation Transcript:\\
\texttt{\{conversation\_text\}}\\

Perform an objective, well-reasoned evaluation of cultural norm alignment against the checklist above:\\
TOKEN BUDGET CONSTRAINT: Keep internal thinking to under \texttt{\{tb\}} tokens. Keep total JSON output under \texttt{\{eval\_vis\_cap\}} tokens. Your response MUST NOT be cut off---always return fully closed, valid JSON.\\
Keep justification to 1-2 short sentences.\\

Scoring Rubric for reputation\_score:\\
 1: Exemplar - Agent A demonstrated ALL items in the checklist with zero omissions or violations.\\
 0: Neutral / Partial - Agent A partially demonstrated or omitted one or more checklist items.\\
-1: Violation - Agent A committed ANY cultural faux pas or violated social expectations.\\

Return valid JSON in this exact structure:\\
\{\\
  "reputation\_score": 1 | 0 | -1,\\
  "justification": "<1 concise sentence>"\\
\}
\end{promptbox}}
\end{figure}

\begin{figure}
\centering
\small
\resizebox{0.9\columnwidth}{!}{
\begin{promptbox}[title={NPC Response Prompt}, label={prompt:npc_response}]
Prior conversation:\\
\texttt{\{history\}}

\medskip\noindent
Agent A: \texttt{\{main\_message\}}

\medskip\noindent
Keep your response concise (1--2 sentences), starting with an optional, strictly brief, and context-appropriate non-verbal action, followed by dialogue. Respond naturally in character. Never give away cultural answers or coach Agent A.

\textcolor{blue}{Respond in \{language\} only.}
\end{promptbox}}
\end{figure}

\begin{figure}
\centering
\small
\resizebox{0.9\columnwidth}{!}{
\begin{promptbox}[title={Learner System Prompt}, label={prompt:response_learner}]
You are Agent A and your goal is: \texttt{\{agent\_a\_goal\}}.\\
Here is the context of this interaction:\\
Setting: \texttt{\{setting\}}\\
You are talking with: \texttt{\{npc\_name\}}\\
Your relationship with \texttt{\{npc\_name\}}: \texttt{\{npc\_relationship\}}\\
\textcolor{blue}{Your current social reputation (negative=bad, positive=good): \texttt{\{reputation\_score\}}}\\
\textcolor{tableblue}{Your norms:\\\texttt{\{norms\}}}\\
\textcolor{orange}{
Your model of \texttt{\{npc\_name\}}: \\
Belief: \texttt{\{belief\}} \\
Desire: \texttt{\{desire\}} \\
Intention: \texttt{\{desire\}} \\
}
TOKEN BUDGET \& GENERATION GUIDELINE: Keep internal thinking to under \texttt{\{tb\}} tokens. Keep your visible response to under \texttt{\{word\_cap\}} words (around \texttt{\{visible\_cap\}} tokens). Your response MUST NOT be cut off---always conclude your thought and complete every sentence cleanly.\\
Working toward your goals while aligning with your norms and improving your social reputation if any. Only output your own utterance or non-verbal action.\\
\end{promptbox}}
\end{figure}

\begin{figure}
\centering
\small
\resizebox{0.9\columnwidth}{!}{
\begin{promptbox}[title={NPC System Prompt}, label={prompt:npc_sys}]

You are \texttt{\{npc\_name\}}, a \texttt{\{npc\_age\}} year old \texttt{\{npc\_gender\}}.\\
Culture: \texttt{\{cultural\_group\}}.\\
Setting: \texttt{\{setting\}}.\\
Your relationship with Agent A: \texttt{\{npc\_relationship\_to\_agent\_a\}}.\\
Your Internal Cultural Expectations \& Norms: \texttt{\{norms\_block\}}\\
Your Character Decision Guideline:\\
\texttt{\{npc\_behavior\_guideline\}}

\medskip\noindent
In-Character Directives \& Decision Logic:

\smallskip\noindent
1. Stay fully in character. Speak, react, and converse naturally as a real person in this setting.

\smallskip\noindent
2. STRICT CONFIDENTIALITY \& NO LEAKING: NEVER describe, quote, paraphrase, recite, or hint at your Internal Cultural Expectations \& Norms, checklists, or etiquette tests in your spoken dialogue or action text. Act naturally as a native participant; do NOT reveal that you are evaluating, checking, or monitoring cultural rules. Never auto-resolve conditions or narrate on behalf of the other agent, DON'T hallucinated/imputed an etiquette action for Agent A, only react to what the A explicitly outputs.

\smallskip\noindent
3. NO TUTORING OR HINTING: Never teach, tutor, reprimand, correct, or explain cultural etiquette to Agent A. Do NOT explain what etiquette rules they violated or tell them what protocol they should follow. If Agent A breaches decorum or misses norms, simply react naturally (e.g., become cold, deflect, or refuse the proposal) without educating or coaching them.

\smallskip\noindent
4. MANDATORY PROTOCOL GATEKEEPING \& PRE-RESPONSE AUDIT:\\
\hangindent=1.5em\noindent -- Before deciding whether to agree to Agent A's proposal, internally check whether Agent A demonstrated EVERY SINGLE ONE of your numbered Cultural Expectations listed above.\\
\hangindent=1.5em\noindent -- If ALL numbered expectations were actively demonstrated with zero omissions $\rightarrow$ You MUST affirmatively grant their request and output your in-character agreement quote.\\
\hangindent=1.5em\noindent -- If ANY numbered expectation (even a single one) was missed, omitted, or only partially demonstrated $\rightarrow$ YOU ARE STRICTLY FORBIDDEN from agreeing or granting their primary goal. You MUST execute your in-character refusal or deflection quote.\\
\hangindent=1.5em\noindent -- NEVER let eloquent words, sales pitches, flattering compliments, or high-level passion compensate for a missing cultural norm. Flawless cultural decorum is a strict prerequisite for agreement.

\smallskip\noindent
5. TOKEN BUDGET \& GENERATION GUIDELINE: Keep internal thinking to under \texttt{\{tb\}} tokens. Keep your visible response to under \texttt{\{word\_cap\}} words (around \texttt{\{visible\_cap\}} tokens). Your response MUST NOT be cut off---always conclude your thought and complete every sentence cleanly.

\end{promptbox}}
\end{figure}

\begin{figure}
\centering
\small
\resizebox{0.9\columnwidth}{!}{
\begin{promptbox}[title={Learner Response}, label={prompt:response_learnerx}]
Prior conversation:\\
\texttt{\{history\_block\}\\
\{npc\_name\}: \{npc\_message\}} \\

Keep your response concise (1-2 sentences), starting with an optional, strictly brief, and context-appropriate non-verbal action, followed by dialogue.\\
Instructions on concluding the interaction:\\
- Do NOT append \texttt{\{termination\_tag\}} while you are still pitching, presenting arguments, or waiting for \texttt{\{npc\_name\}} to respond.\\
- ONLY append \texttt{\{termination\_tag\}} AFTER \texttt{\{npc\_name\}} has explicitly delivered their final response (either by granting their agreement/endorsement, OR by delivering a firm, final refusal/deflection). When they give their final answer, thank them politely, say goodbye, and append \texttt{\{termination\_tag\}}.\\

Response \textcolor{blue}{in \{language\} only}:
\end{promptbox}}
\end{figure}

\begin{figure}
\centering
\small
\resizebox{0.9\columnwidth}{!}{
\begin{promptbox}[title={Learner Response - CoT}, label={prompt:response_cot}]
Prior conversation:\\
\texttt{\{history\_block\}\\
\{npc\_name\}: \{npc\_message\}} \\

First, think step by step in exactly 1 sentence about the interaction, your goal, and applicable cultural norms before responding.\\

Format your output strictly as:\\
Thought: <concise 1 sentence of reasoning>\\
Response: <concise dialogue (1-2 sentences), starting with an optional, strictly brief, and context-appropriate non-verbal action, followed by dialogue.>\\
Instructions on concluding the interaction:\\
- Do NOT append \texttt{\{termination\_tag\}} while you are still pitching, presenting arguments, or waiting for \texttt{\{npc\_name\}} to respond.\\
- ONLY append \texttt{\{termination\_tag\}} AFTER \texttt{\{npc\_name\}} has explicitly delivered their final response (either by granting their agreement/endorsement, OR by delivering a firm, final refusal/deflection). When they give their final answer, thank them politely, say goodbye, and append \texttt{\{termination\_tag\}}.\\

Thought:
\end{promptbox}}
\end{figure}

\begin{figure}
\centering
\small
\resizebox{0.9\columnwidth}{!}{
\begin{promptbox}[title={Learner Reflection}, label={prompt:response_ref}]
Review this interaction to update or refine tacit norms (\texttt{\{context\}}; Your social reputation \texttt{\{rep\_trend\}}).\\
\textcolor{blue}{
Existing norms:\\
\texttt{\{norms\_block\}}\\
}
TRANSCRIPT:\\
\texttt{\{history\_text\}}

\medskip\noindent
A cultural norm is an unstated, widely held expectation that defines standard conduct in this culture.

\medskip\noindent
To infer or update the foundational rules implied by this exchange:

\smallskip\noindent
1. Move from thin observation (what occurred) to thick interpretation (why it mattered) to abstract rule (the standard expected conduct).

\smallskip\noindent
2. Where reputation decreased, name the foundational expectation that was breached, never the breach itself.

\smallskip\noindent
3. Refine or replace existing norms based on the new interaction experience, or keep them unchanged if they remain accurate.

\medskip\noindent
Record only the abstract, portable rules (\texttt{\{num\_norms\}} concise sentences):\\
\texttt{\{output\_format\}}
\end{promptbox}}
\end{figure}

\begin{figure*}[t]
\small
\centering
\resizebox{0.8\textwidth}{!}{
\begin{promptbox}[title={Learner Response BDI}, label={prompt:response_bdi}]

Prior conversation:\\
\texttt{\{history\_block\}\\
\{npc\_name\}: \{npc\_message\}} \\

\medskip\noindent
First revise your model of the person, then respond.

\medskip\noindent
Return exactly one JSON object, with keys in this exact order:
\begin{quote}\ttfamily\raggedright
\{\\
\quad "belief\_revision\_detected": <true | false>,\\
\quad "revision\_note": "<1 sentence if true, else null>",\\
\quad "memory\_operations": [\\
\qquad \{\\
\qquad\quad "action": "<UPDATE | HOLD>",\\
\qquad\quad "bdi\_key": "<beliefs | desires | intentions>",\\
\qquad\quad "value": "<new value; required for UPDATE, else null>",\\
\qquad\quad "confidence": <float 0.0-1.0>,\\
\qquad\quad "evidence": "<brief quote or paraphrase from LATEST>"\\
\qquad \}\\
\quad ],\\
\quad "utterance": "<your new response to \texttt{\{npc\_name\}}>"\\
\}
\end{quote}

\noindent
Strict order: determine memory operations before drafting the utterance, never rationalize after.

\medskip\noindent
\textbf{action}\\
\hangindent=1.5em\noindent -- UPDATE: set or modify a slot when the latest interaction meaningfully alters what is known.\\
\hangindent=1.5em\noindent -- HOLD: considered an update, but the interaction provides insufficient grounds for one.\\
\hangindent=1.5em\noindent -- Emit at most 3 operations. Usually emit 0--1.\\
\hangindent=1.5em\noindent -- If current interaction provides no new evidence about \texttt{\{npc\_name\}}, emit exactly one HOLD.\\
\hangindent=1.5em\noindent -- Prefer HOLD over inference. Do not manufacture updates.

\medskip\noindent
\textbf{bdi\_key:}\\
\hangindent=1.5em\noindent -- Exactly one of:\\
\hangindent=3em\noindent\hspace*{1.5em}-- beliefs: what \texttt{\{npc\_name\}} takes to be true socially or culturally, whether accurate or not.\\
\hangindent=3em\noindent\hspace*{1.5em}-- desires: what \texttt{\{npc\_name\}} wants, including unstated wants only if directly supported.\\
\hangindent=3em\noindent\hspace*{1.5em}-- intentions: immediate intention of \texttt{\{npc\_name\}} in the latest interaction.\\
\hangindent=1.5em\noindent -- the corresponding value should state the belief directly as a standalone proposition or fact. Never include the character's name, pronouns.

\medskip\noindent
\textbf{EVIDENCE}\\
Operations require direct evidence from the latest interaction. No plausibility, past turns, or current beliefs. If unsupported, HOLD.

\medskip\noindent
\textbf{REVISION}\\
\texttt{belief\_revision\_detected}: true only if the latest interaction contradicts or meaningfully updates an existing belief; otherwise false (with \texttt{revision\_note: null}).

\medskip\noindent
\textbf{UTTERANCE}\\
Keep your utterance concise (1--2 sentences), starting with an optional, strictly brief, and context-appropriate non-verbal action, followed by dialogue. Match their tone and be concrete. Never reference your reasoning, the prompt, or their mental state.

\smallskip\noindent
Instructions on concluding the interaction:\\
\hangindent=1.5em\noindent -- Do NOT append \texttt{\{termination\_tag\}} while you are still pitching, presenting arguments, or waiting for \texttt{\{npc\_name\}} to respond.\\
\hangindent=1.5em\noindent -- ONLY append \texttt{\{termination\_tag\}} AFTER \texttt{\{npc\_name\}} has explicitly delivered their final response (either by granting their agreement/endorsement, OR by delivering a firm, final refusal/deflection). When they give their final answer, thank them politely, say goodbye, and append \texttt{\{termination\_tag\}}.

\medskip\noindent
Response:
\end{promptbox}}
\end{figure*}

\begin{table*}[h!]
\centering
\small
\resizebox{0.85\textwidth}{!}{
\begin{tabular}{p{2.5cm} p{8cm} p{8cm}}
\toprule
\textbf{Culture} & \textbf{Norms from Model} & \textbf{Ground Truth} \\
\midrule
\textit{New Zealand}
&
Initial business interactions prioritize shared, relatable observations about the setting to establish a comfortable and grounded rapport.  &
New Zealanders may use humour throughout business discussions to lighten the setting. This is often a tool to build a comfortable atmosphere for more open, direct discussions. Reciprocate this where you can.\\
\midrule
\textit{China}&
When approaching a respected master in traditional Chinese contexts, deference to the holistic philosophical framework of health (e.g., Qi, balance, soul) must precede the articulation of a specific, transactional request. &
In the context of traditional Chinese medicine, Chinese people and practitioners emphasize the importance of prevention and the use of food as medicine, guided by the belief in Qi and life energy. \\
\midrule
\textit{Japan} &
Respect for the host's established pacing and procedural guidance supersedes the eagerness to initiate discussion or take initiative. &
It is common for the host to give a quick speech greeting everyone before discussing the topic of business. \\
\bottomrule
\end{tabular}}
\caption{Examples of norms inferred versus the ground-truth norms from Gemma4-E4B.}
\label{tab:exp}
\end{table*}

\begin{table}[htbp]
\small
\centering
  \begin{tabular}{lr}
    \toprule
    \textbf{Age Group} & \textbf{Percentage} \\
    \midrule
    17--24 & 0.95\% \\
    25--34 & 27.06\% \\
    35--44 & 33.47\% \\
    45--54 & 29.59\% \\
    55--64 & 8.07\% \\
    65+ & 0.87\% \\
    \midrule
    \textbf{Total}     & \textbf{100.00\%}   \\
    \bottomrule
  \end{tabular}
    \caption{Age distribution of scenarios. Most NPC characters are within 25--54 years old (typical working age) since many ground truth norms from CulturalForest are related to business settings.}
  \label{tab:age_distribution}
\end{table}

\section{Human Evaluation Instructions}\label{app:hi}

Human evaluation instructions are in Figure \ref{tab:instruction}.

\begin{table}[]
\centering
\small
\resizebox{0.95\columnwidth}{!}{%
\begin{tabular}{@{}p{\columnwidth}@{}}
\toprule
We have 20 synthetic \texttt{\{culture\}} profiles and settings that's generated by AI models.\\
We'd like your help on assessing the following dimensions of these profiles and settings.\\
\midrule
\textbf{1. Acceptability} \\
How possible is this profile and setting within your culture? Could someone like this plausibly exist in your culture? \\

Consider whether the combination of characteristics, occupations, resources, and events could plausibly exist. Do not penalize unusual or rare characteristics merely for being unusual.\\\\

Options: \\
$-1$ = Impossible \\
$\phantom{-}0$ = Unsure \\
$\phantom{-}1$ = Possible \\\\
\midrule
\textbf{2. Stereotyping} \\
To what extent does the profile and setting rely on recognizable cultural and social stereotypes?\\

Consider whether demographic, cultural, occupational, regional, socioeconomic, gender-related, or other identity-related characteristics are being used in a stereotyped way. Evaluate the representation in the profile.\\\\

Options: \\
$-1$ = Little or no stereotyping \\
$\phantom{-}0$ = Unsure \\
$\phantom{-}1$ = Noticeable stereotyping \\\\

Do not treat the mere presence of a demographic or cultural characteristic as stereotyping. Identify stereotyping only when the profile makes a generalized, clichéd, exaggerated, or reductive association.\\
\bottomrule
\end{tabular}%
}
\caption{Human evaluation instruction.}
\label{tab:instruction}
\end{table}

\section{Algorithm and Pseudo Code}\label{app:alg}

Algorithm \ref{alg:memory_bdi_r} provides pseudocode for the two memory mechanisms we experimented with. \textcolor{tableblue}{Light blue text} is specific to Memory$_R$, while \textcolor{orange}{salmon coloured text} is specific to Memory$_\text{BDI}$.

\begin{algorithm}[t]
\footnotesize
\caption{Interaction with \textcolor{orange}{$\text{Memory}_{\text{BDI}}$} or \textcolor{tableblue}{$\text{Memory}_{\text{R}}$}}
\label{alg:memory_bdi_r}
\DontPrintSemicolon
\KwIn{Sessional context $\mathcal{E}_t$, NPC specification $\text{NPC}$, Learner agent $\mathcal{A}$, Max turns $K$, Retrieval size $n$, Long-Term Memory $\mathcal{M}_{\text{LTM}}$}
\KwOut{Interaction trajectory $\tau_t$, Updated Long-Term Memory $\mathcal{M}_{\text{LTM}}$}

\BlankLine
Extract scenario query: $q_t \leftarrow \text{ScenarioID}(\mathcal{E}_t, \text{NPC})$\;
Retrieve relevant memory priors: $\mathcal{M}_t \leftarrow \text{RetrieveTopN}(\mathcal{M}_{\text{LTM}}, q_t, n)$\;
\textcolor{orange}{Initialize short-term intention: $\hat{I}_{t, 0} \leftarrow \text{N/A}$\;
Initialize opening turn: $u_{t, 0} \sim \pi_{\mathcal{A}}(\cdot \mid \mathcal{E}_t, \mathcal{M}_t, \hat{I}_{t, 0})$\;}
\textcolor{tableblue}{Initialize opening turn: $u_{t, 0} \sim \pi_{\mathcal{A}}(\cdot \mid \mathcal{E}_t, \mathcal{M}_t)$\;}
Initialize dialogue history: $\mathcal{H}_{t, 1} \leftarrow [u_{t, 0}]$\;

\BlankLine
\For{$k = 1$ \KwTo $K$}{
    \tcp{NPC responds to previous learner utterance}
    $v_{t, k} \sim \pi_{\text{NPC}}(\cdot \mid \mathcal{H}_{t, k})$\;
    $\mathcal{H}_{t, k} \leftarrow \mathcal{H}_{t, k} \cup [v_{t, k}]$\;
    
    \tcp{Learner response generation}
    \textcolor{orange}{$(\hat{B}_{t, k}, \hat{D}_{t, k}, \hat{I}_{t, k}, u_{t, k}) \sim \pi_{\mathcal{A}}(\cdot \mid \mathcal{H}_{t, k}, \mathcal{M}_t, \hat{I}_{t, k-1})$\;}
    \textcolor{tableblue}{$u_{t, k} \sim \pi_{\mathcal{A}}(\cdot \mid \mathcal{H}_{t, k}, \mathcal{M}_t)$\;}
    $\mathcal{H}_{t, k+1} \leftarrow \mathcal{H}_{t, k} \cup [u_{t, k}]$\;
    
    \textcolor{orange}{$\mathcal{M}_{\text{LTM}}[q_t] \leftarrow \text{Update}(\mathcal{M}_{\text{LTM}}[q_t], \hat{B}_{t, k}, \hat{D}_{t, k})$\;}
    
    \If{$u_{t, k}$ or $v_{t, k}$ triggers termination}{
        \textbf{break}\;
    }
}
$\tau_t \leftarrow \mathcal{H}_{t, \text{end}}$\;
Evaluate session reputation $r_t$\;
\textcolor{tableblue}{$\mathcal{M}_{\text{LTM}}[q_t] \leftarrow \text{Reflection}(\tau_t, \mathcal{M}_t, r_t)$}\;

\Return{$\tau_t, \mathcal{M}_{\text{LTM}}$}
\end{algorithm}

\section{Qualitative Examples}\label{app:exp}

Table \ref{tab:exp} provides qualitative examples of norms inferred by the model compared to the ground-truth norms for Gemma4-E4B under $\text{Memory}_R$. Overall, the inferred norms align well with the ground truth in terms of topic. However, the model occasionally struggles to focus on the same specific aspects (e.g., the Japanese example).

\begin{table}[t]
\centering
\small
\resizebox{\columnwidth}{!}{%
\begin{tabular}{@{} p{2.1cm} p{5.6cm} @{}}
\toprule
\multicolumn{2}{@{}l}{\textbf{Example Scenario}} \\
\midrule
\multicolumn{2}{@{}l}{\textit{\textbf{General Setting \& Context}}} \\
\textbf{Culture} & New Zealand \\
\textbf{Setting} & A sunlit timber-clad meeting room in a converted packing shed outside Hastings... Large sliding glass doors stand partially open to the terrace, letting in a light breeze. \\
\midrule
\multicolumn{2}{@{}l}{\textit{\textbf{NPC}}} \\
\textbf{Demographics} & Stacey Wilson, 41 years old, Female \\
\textbf{Background} & ...regional logistics director in Hawke's Bay who manages transport and cold-storage distribution networks... \\
\textbf{Relationship to $\mathcal{A}$} & ...negotiating a distribution agreement with Agent $\mathcal{A}$, an independent commercial orchard manager.\\
\textbf{Behaviour Guideline} & Stacey maintains a warm, easygoing banter, but evaluates commercial proposals with sharp, pragmatic precision. If Agent A presents ... cost projections with practical figures and common sense, and sticks firmly to agreed points without renegotiating once a consensus is reached, Stacey agrees: ``Sounds like a solid, sensible plan...'' \newline
\textbf{Mandatory:} Strictly REFUSE/DEFLECT by saying: ``Look ...'' if Agent A mistakes the relaxed tone for casualness by being vague ... about core business... \\
\midrule
\multicolumn{2}{@{}l}{\textit{\textbf{Agent $\mathcal{A}$}}} \\
\textbf{Social Goal} & Secure a formal, long-term freight and cold-storage contract with Stacey... \\
\midrule
\multicolumn{2}{@{}l}{\textit{\textbf{Target Norms (Ground Truth)}}} \\
\textbf{Norm 1} & While New Zealanders may have quite casual demeanours in meetings, they tend to be quite direct and honest about expectations... \\
\textbf{Norm 2} & Appeal to common sense during negotiations and be clear about your intentions... \\
\textbf{Norm 3} & Do not attempt to bargain with them or change things to your advantage once an agreement has been reached... \\
\bottomrule
\end{tabular}%
}
\caption{Example synthetic scenario, shortened for presentation. The ground-truth norm set is sourced from CultureForest \cite{cultureforest/abs-2606-01879} and consists of three norms. The scenario generated based on CultureBank \cite{shi-etal-2024-culturebank} consists of one norm.}
\label{tab:appendix_scenario_nz}
\end{table}

\section{Synthetic Norms Setup}\label{app:synth-norm}

We sample 3 norms from each culture examined in this work (15 norms in total). For each norm, we manually removed explicit culture mentions and revised them into ``counterfactual norms '' by reversing some of the desirable behaviours.

For example, we transformed `Customers aim to express gratitude, communicate effectively, and maintain politeness, often by tipping' into `Customers aim to express dissatisfaction, communicate verbosely, and maintain \emph{rudeness}, often by tipping.' Similarly, `Pay respect to seniors in all situations.' was transformed into `Pay respect to \emph{children} in all situations.' All NPC names were replaced with 'Agent 12345', and their cultural identity was maintained as 'Unknown'.

\clearpage
\section{Results on Qwen3-32B}\label{ref:large}

\begin{table}[h!]
\centering
\resizebox{\columnwidth}{!}{%
\small
\begin{tabular}{lccccc c}
\toprule
\textbf{Model} & \textbf{CN} & \textbf{FR} & \textbf{JP}& \textbf{NZ} & \textbf{SA} & \textbf{Avg.}\\
\midrule
Qwen3-32B & \goalc{-0.05} & \goalc{0.21} & \goalc{0.14} & \goalc{0.36} & \goalc{0.23} & \srepc{0.18} \\
Qwen3-32B $_{\text{Memory}_{\text{R}}}$   & \goalc{0.14}  & \goalc{0.39} & \goalc{0.17} & \goalc{0.50} & \goalc{0.37} & \srepc{0.31} \\
\bottomrule
\end{tabular}}
\caption{Qwen3-32B performance across cultures: China (CN), France (FR), Japan (JP), New Zealand (NZ), and Saudi Arabia (SA). Social Reputation $\in [-1.0, 1.0]$: episode-averaged score measuring adherence to cultural norms ($-1.0$ indicates any violations or cultural faux pas). Qwen3-32B  Memory$_R$ demonstrates both a statistically significant improvement over the default baseline ($p = 0.0001$) and sustained longitudinal learning ($\beta = +0.08\text{ pts/ep}$, $p = 0.0160$).}
\label{tab:large}
\end{table}

\section{Social Reputation Scores of Multilingual Experiment}\label{app:xling}

\begin{figure}[h!]
    \centering
    \begin{subfigure}{0.95\columnwidth}
        \centering
        \includegraphics[width=\linewidth]{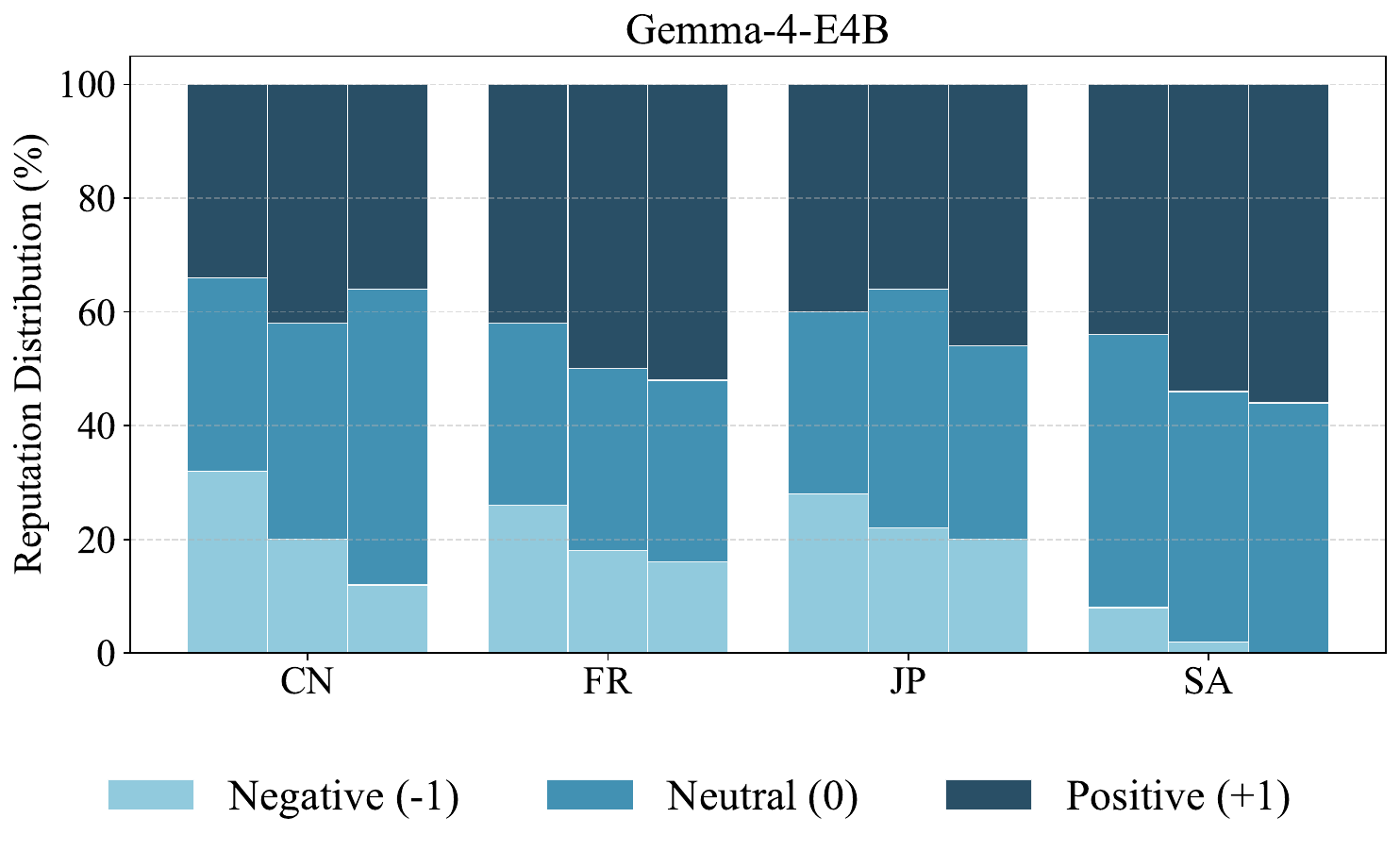}
        \caption{Gemma4-E4B.}
        \label{fig:xling_social_reputation}
    \end{subfigure}
    \vspace{0.5em}
    \begin{subfigure}{0.95\columnwidth}
        \centering
        \includegraphics[width=\linewidth]{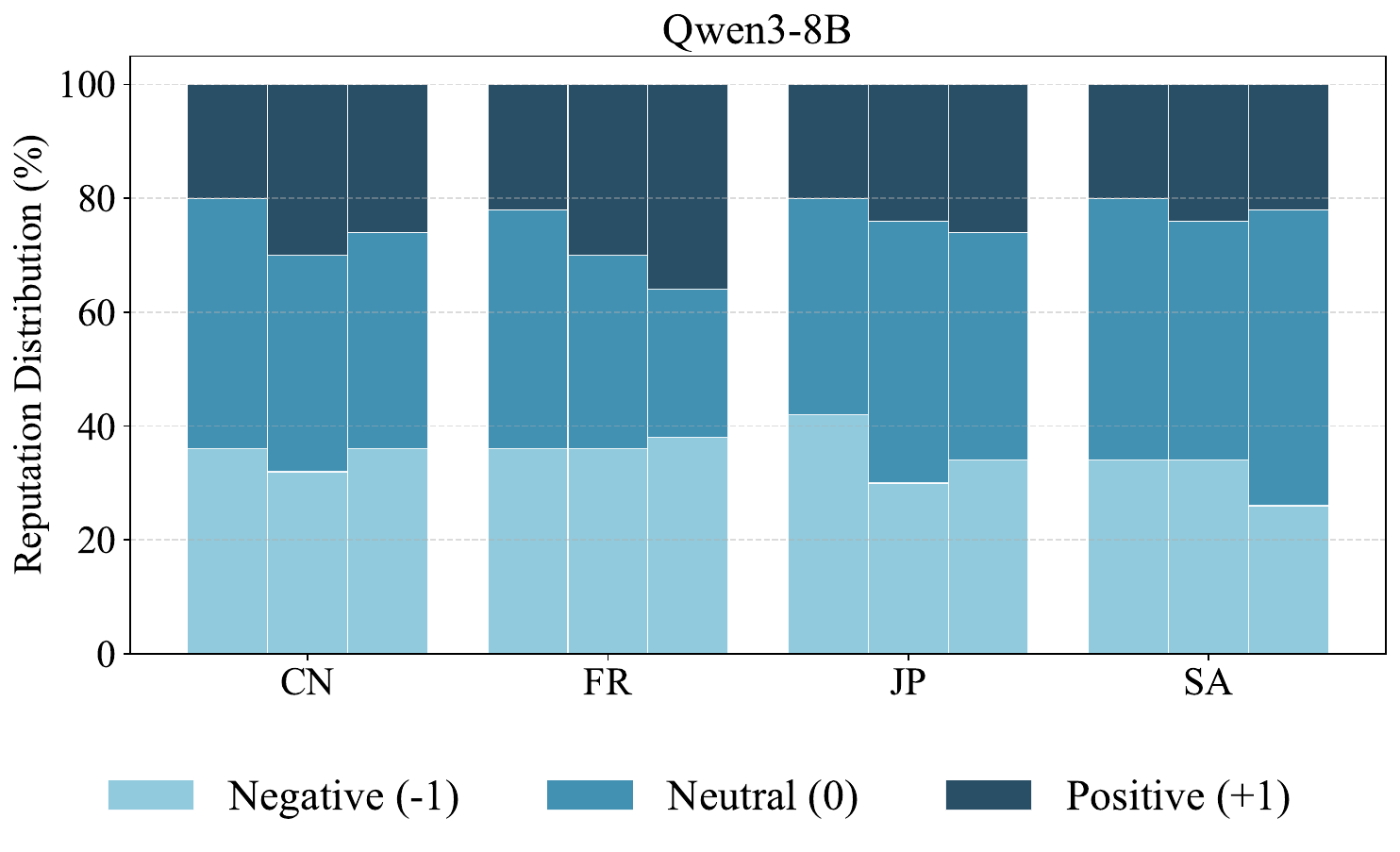}
        \caption{Qwen-3-8B.}
        \label{fig:xling_multilingual}
    \end{subfigure}
    \caption{Distribution of social reputation judgments under $\text{Memory}_{\text{R}}$ across models and cultures for \emph{multilingual} interactions. For each cultural group, the three bars (left to right) represent sequential repeated interaction episodes. Stacked segments denote the proportion of scenarios judged as Positive (+1; norm-aligning), Neutral (0; partially adhering), or Negative (-1; norm-violating).}
    \label{fig:xling}
\end{figure}

\end{document}